\documentclass{article}

\PassOptionsToPackage{numbers,compress}{natbib}
\usepackage[preprint]{arxiv_preprint}
\usepackage[utf8]{inputenc}
\usepackage[T1]{fontenc}
\usepackage{hyperref}
\usepackage{url}
\usepackage{microtype}
\usepackage{graphicx}
\usepackage{subcaption}
\usepackage{enumitem}
\usepackage{booktabs}
\usepackage{multirow}
\usepackage{xcolor}
\usepackage{colortbl}
\usepackage{balance}
\usepackage{placeins}
\usepackage{tabularx}
\usepackage{amsmath}
\usepackage{amsfonts}
\usepackage{amssymb}
\usepackage{mathtools}
\usepackage{array}
\usepackage[capitalize,noabbrev]{cleveref}
\usepackage{algorithm}
\usepackage{algorithmic}

\newcolumntype{P}[1]{>{\raggedright\arraybackslash}p{#1}}
\newcolumntype{Y}{>{\centering\arraybackslash}X}
\definecolor{referitsectionbg}{RGB}{241,246,250}
\definecolor{referitreferencebg}{RGB}{246,246,246}
\newcommand{\ftyes}{\textcolor{green!55!black}{$\checkmark$}}
\newcommand{\ftno}{\textcolor{red!70!black}{$\times$}}

\title{SSR3D-LLM: Structured Spatial Reasoning via Latent Steps for Fine-Grained Grounding in Unified 3D-LLMs}
\author{
  \textbf{Jiawei Li}$^{1}$ \quad
  \textbf{Ziyi Liu}$^{2}$ \quad
  \textbf{Weijie Shi}$^{1}$ \\
  \textbf{Long Chen}$^{1}$ \quad
  \textbf{Jiajie Xu}$^{2}$ \quad
  \textbf{Xiaofang Zhou}$^{1}$ \\
  \normalfont $^{1}$The Hong Kong University of Science and Technology \\
  $^{2}$Soochow University
}

\begin{document}

\maketitle

\begin{abstract}
3D object grounding localizes referred objects in a 3D scene from natural language.
Unified instance-centric 3D-LLMs aim to solve grounding together with dialog, QA, and captioning, yet many rely on a single pointer-style grounding decision that compresses a relational instruction into one selection.
This is brittle for fine-grained queries where multiple same-class candidates must be ruled out by context objects and spatial relations.
We propose \emph{Structured Spatial Reasoning 3D-LLM} (SSR3D-LLM), a structured grounding interface for unified 3D-LLMs.
Given fixed Mask3D object proposals, the LLM writes a sequence of latent spatial reasoning steps and memory tokens from the query, and a geometry-aware scorer reads these latent steps in order to refine candidate rankings step by step with step-length masking.
The latent steps are learned from standard benchmark target supervision with auxiliary referential-cue supervision during training, while inference uses only the input query and Mask3D proposals.
Across ReferIt3D, ScanRefer, and Multi3DRef, SSR3D-LLM achieves the strongest results among unified 3D-LLM baselines, with substantial gains over the single-pointer \emph{QPG} baseline on fine-grained grounding and consistent improvements over prior unified 3D-LLMs, while preserving the default language-task route.
\end{abstract}

\section{Introduction}
\label{sec:introduction}
When we instruct a robot to find an object in a 3D scene, many real-world instructions are fine-grained and compositional, specifying not only object category and attributes, but also spatial position and relations to other context objects, e.g., ``find the white chair to the right of the brown sofa in the center of the room''.
Here, the target object is the referent to be localized, while context objects are the other mentioned objects that help disambiguate it, such as ``the brown sofa''.
The robot must parse the linguistic structure and combine it with geometric cues from the 3D scene to precisely localize the target instance.
This is the 3D object grounding task, a core capability for embodied AI and spatial intelligence that enables downstream manipulation, navigation, and situated question answering~\cite{duan2022survey,liu2024survey}.

\begin{figure*}[t]
  \centering
  \includegraphics[width=\textwidth]{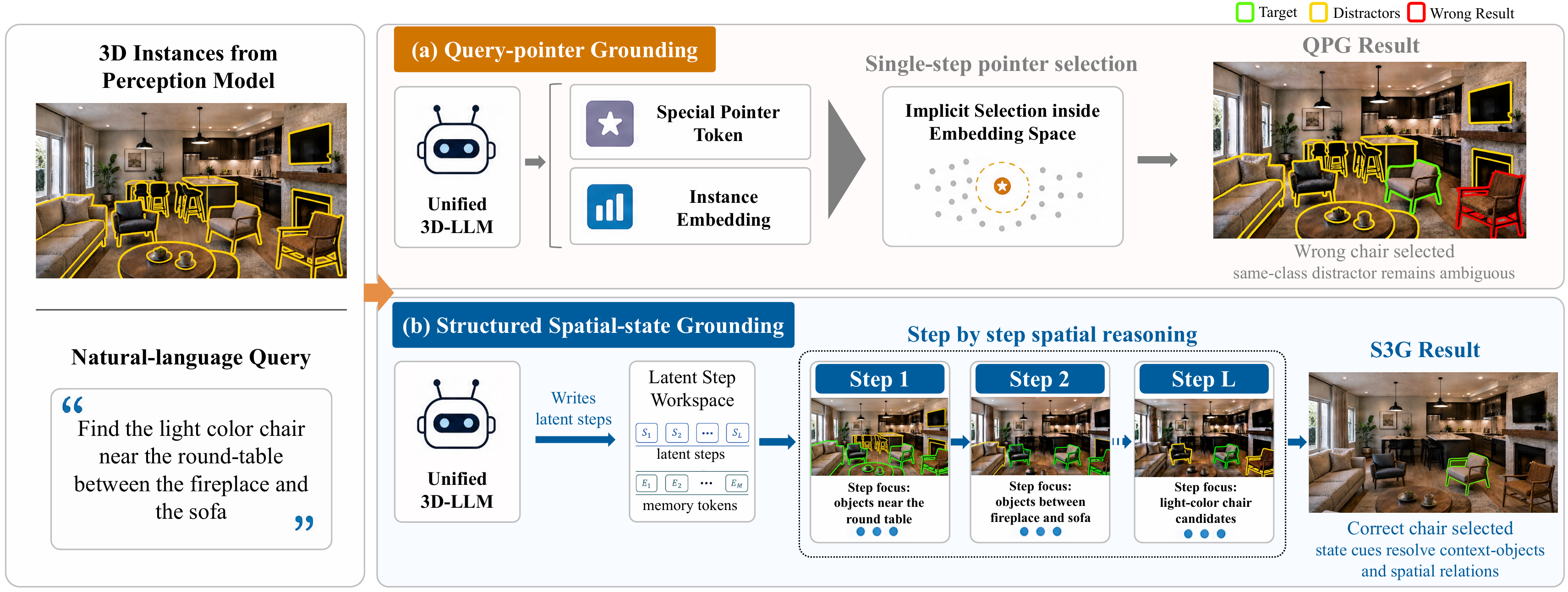}
  \caption{\textbf{QPG vs. S3G.} Fine-grained 3D grounding often requires ruling out candidates through context objects and spatial relations. QPG makes one pointer-style selection, while S3G writes latent spatial reasoning steps and refines candidate rankings step by step.}
  \label{fig:qpg_s3g_paradigms}
  \vspace{-2em}
\end{figure*}

Early 3D object grounding methods mainly learned supervised cross-modal matching on human-annotated datasets such as ScanRefer and ReferIt3D~\cite{chen2020scanrefer,achlioptas2020referit_3d,dai2017scannet}.
They can work well on benchmark expressions, but their language side is usually task-specific: it is less suited to open-form instructions and does not naturally share a model with dialog, question answering, or captioning.
Recent foundation models that jointly process text and 3D scenes address this language-interface limitation by building \emph{3D-LLMs} that support language generation and 3D object grounding in a single model~\cite{chen2024grounded,huang2024chat,yu2023comprehensive,azuma2022scanqa,jia2024sceneverse,yang2024grand,li2024dense}.
Representative grounding-capable unified 3D-LLMs, such as Grounded 3D-LLM~\cite{chen2024grounded} and Chat-Scene~\cite{huang2024chat}, extend instruction-tuned LLMs with 3D object representations so the same model can both generate language and select referred objects.
Because grounding must share the model with dialog, question answering, and captioning, it is often exposed through a compact output interface rather than a task-specific grounding architecture.
In Grounded 3D-LLM, this appears as \emph{Query-Pointer Grounding} (\emph{QPG})~\cite{fu2025scene}: a single query/pointer representation is used to select one candidate object representation.
Related unified systems use similarly compact object-id or pointer-style readouts, forming a broader family of single-readout grounding interfaces.
As illustrated in Figure~\ref{fig:qpg_s3g_paradigms}~(a), this makes grounding compatible with the LLM's token-level training interface, but it also gives fine-grained spatial grounding a narrow readout.

However, this design is not sufficient to solve fine-grained compositional grounding due to three challenges.
(i) Grounding becomes a single selection where target category, attributes, context objects, and spatial relations all have to be reflected in one pointer-style decision.
This leaves no intermediate workspace for progressively checking context objects or ruling out same-class distractors, which is exactly what fine-grained spatial queries often require.
(ii) Standard final-target supervision only says which object is correct.
It does not directly supervise how the model should use context objects and relations when comparing several plausible candidates, so a single-step selector can rely on shortcuts that work on easy expressions but fail on queries requiring ordered spatial disambiguation.
(iii) When fine-grained grounding fails, the error may come from parsing the target, choosing the wrong context object, or applying the wrong spatial relation, yet a single pointer output gives little insight into which part failed.

\textbf{Our method}. To address these challenges, we propose \emph{Structured Spatial-State Grounding (S3G)}, shown in Figure~\ref{fig:qpg_s3g_paradigms}~(b), and instantiate it as \emph{Structured Spatial Reasoning 3D-LLM (SSR3D-LLM)}.

First, to better handle fine-grained queries with multiple context objects and spatial relations, S3G adds a sequence of latent spatial reasoning steps inside the 3D-LLM, giving the grounding branch a lightweight structured interface instead of squeezing all query cues into a single pointer token.
Each step is a latent state read from a reserved step marker, and memory tokens provide shared query context across steps.
The steps are not generated text, so the model keeps a compact latent grounding interface without exposing an explicit chain-of-thought output.

Second, to ensure the model actually uses these latent steps, we supervise the reserved step markers during training with auxiliary referential cues, on top of the same standard query-target supervision used for grounding.
When intermediate cues are not provided by a benchmark, we derive them offline from the training queries and annotations, so the S3G pathway remains supervised without changing the test-time input.
At inference, the model receives only the input query and fixed Mask3D proposals.

Third, to make the procedure stable and easy to inspect, a geometry-aware scorer refines candidate rankings step by step over the fixed Mask3D proposal set.
Step-length masking freezes updates after the effective steps are done, and the resulting per-step rankings form a compact non-textual trace that helps localize failures.
SSR3D-LLM keeps the upstream point-cloud-derived proposal pipeline and the LLM backbone family in place, and changes how grounding information is represented and consumed.
Finally, because a 3D-LLM must remain a capable language model, we confine grounding-specific parameter updates to lightweight adapters and grounding modules to minimize interference with default language behavior.
Code and data will be released.

Our main contributions are:
\begin{itemize}[leftmargin=1.2em, nosep]
  \item We introduce SSR3D-LLM, a unified 3D-LLM framework that keeps the default dialog/QA/captioning pathway while adding a structured grounding route for fine-grained object localization.
  \item We propose a structured grounding method S3G, which replaces single-pointer QPG with latent spatial reasoning steps learned from training-time referential cues, a geometry-aware candidate scorer, step-length masking, and an inspectable score trace.
  \item We evaluate on ReferIt3D, ScanRefer, and Multi3DRef: our SSR3D-LLM outperforms Grounded 3D-LLM and Chat-Scene across fine-grained candidate selection, box localization, and set-level grounding within the unified 3D-LLM family, while also achieving lower grounding-readout latency and higher throughput after proposal extraction.
\end{itemize}

\section{Related Work}

\noindent\textbf{Early and specialized 3D grounding systems} treat grounding as a dedicated cross-modal matching problem over 3D scenes and referring expressions~\cite{chen2020scanrefer,achlioptas2020referit_3d}.
Many later models add explicit geometric priors, such as distances, directions, object relations, and candidate comparison, to distinguish similar objects in cluttered scenes~\cite{roh2021languagerefer,Zhao_2021_ICCV,luo2022_3dsps,He2021TransRefer3D,zhang2023multi3drefer}.
Representative systems such as MiKASA~\cite{chang2024mikasa}, ViL3DRel~\cite{chen2022language}, and MVT~\cite{huang2022multi} show that geometry-aware interaction is important for fine-grained 3D reference resolution.
These designs motivate our geometry-aware candidate scorer, but they are usually task-specific rather than unified language models.

\noindent\textbf{Large-model and agentic 3D grounding} methods use stronger language and visual priors through prompting, rendered views, external 2D detectors, or verification loops.
Zero-shot or training-free systems such as ZSVG3D~\cite{Yuan2023Visual}, SeeGround~\cite{li2025seeground}, LaSP~\cite{mi2025lasp}, and SORT3D~\cite{zantout2025sort3d} treat grounding as a dedicated inference pipeline with rendered or multi-view evidence.
Reasoning-centric systems such as LLM-Grounder~\cite{yang2023llm}, VLM-Grounder~\cite{xu2024vlmgrounder}, ViewRefer~\cite{guo2023viewrefer}, ViewInfer3D~\cite{Geng2024ViewInfer3D}, Transcrib3D~\cite{fang2024transcrib3d}, SPAZER~\cite{jin2025spazer}, and GPT4Scene~\cite{qi2025gpt4scene} use external planning, 2D/multi-view foundation-model prompting, or agentic candidate verification to improve grounding.
These methods provide useful broader context and often strong performance, but they operate under different system assumptions from our instance-centric unified 3D-LLM setting.

\noindent\textbf{Unified 3D-LLMs for grounding} extend instruction-tuned LLMs with instance-centric 3D representations so a single model can both generate language and ground a referred target object in a 3D scene~\cite{chen2024grounded,huang2024chat,hong20233d,jia2024sceneverse,yang2024grand,li2024dense}.
Grounded 3D-LLM~\cite{chen2024grounded} and Chat-Scene~\cite{huang2024chat}, among others, realize grounding by having the model output a discrete referent token or object identifier that selects one object representation.
This pointer-style interface (\emph{QPG})~\cite{fu2025scene} keeps grounding compatible with token prediction, but it gives fine-grained spatial reasoning only a single-step readout.
SSR3D-LLM keeps the same unified-model premise and point-cloud-derived candidate setting, but replaces the single pointer route with compact latent spatial reasoning steps for step-wise candidate ranking.

\noindent\textbf{Intermediate supervision and capability preservation} such as referential order or context-object chains can make fine-grained grounding easier to guide and inspect~\cite{abdelrahman2023cot3dref,wu2025data,abdelreheem2024scanents3d}.
Our \emph{S3G} uses such cues to shape an ordered internal state, while inference still uses only the input query and 3D candidates.
At the same time, grounding-specific supervision can interfere with the language-generation behavior of multimodal LLMs.
LISA~\cite{lai2024lisa} ties a language token embedding to a segmentation control vector, while STAMP~\cite{liu2025betterstrongerfastertackling} confines dense supervision to the visual side to better preserve conversational capability.
Following this concern, we keep the default language-generation pathway intact and confine grounding-specific training signals to lightweight adaptation and grounding modules.
Surveys on spatial reasoning in multimodal LLMs~\cite{liu2025spatial} similarly highlight the value of separating language generation from geometric decision-making.

\section{Preliminaries}
\label{sec:prelim}

\noindent\textbf{Data and Inputs.}
We use an instance-centric 3D scene \(S\), a referring query \(u\), and \(N\) candidate objects or proposals indexed by \(i\), from official ReferIt3D sets or Mask3D proposals after benchmark matching.
For proposal \(i\), \(\mathbf{f}_i,\mathbf{b}_i,\mathbf{a}_i,\mathbf{c}_i\) denote its pooled Mask3D feature, rotated-box geometry, optional DINOv2 multi-view appearance, and proposal-side predicted-class label embedding.

\noindent\textbf{Candidate-Set Formulation.}
Grounding ranks \(\{1,\dots,N\}\) with logits \(\boldsymbol{\ell}\in\mathbb{R}^{N}\).
The supervised answer is a target index \(i^\star\) for single-target localization (Nr3D and ScanRefer-style tasks) or a target set \(Y^\star\subseteq\{1,\dots,N\}\) for set-level grounding such as Multi3DRef.

\noindent\textbf{Learning Setting.}
We study supervised instance-level 3D grounding in a 3D-LLM that also supports general language generation.
In S3G, \(D\) is the scorer dimension, \(K\) the latent-step budget, and \(M\) the memory-token count; \texttt{\textless geom\textgreater} activates grounding, with reserved step and memory tokens \(\tau^{\mathrm{step}}_k\) and \(\tau^{\mathrm{mem}}_j\).
Auxiliary referential cues supervise latent steps during training only and are not part of the inference input.

\noindent\textbf{Problem Definition.}
In grounding mode, SSR3D-LLM predicts \(\hat{i}=\arg\max_i \boldsymbol{\ell}_i\) for single-target evaluation, or thresholds/ranks logits under the benchmark protocol to obtain \(\hat{Y}\subseteq\{1,\dots,N\}\) for set-level evaluation.
All metrics are computed after the candidate matching or restriction required by each benchmark.
\vspace{-1.0em}

\section{Method}
\label{sec:method}
\begin{figure*}[t]
  \centering
  \includegraphics[width=\textwidth]{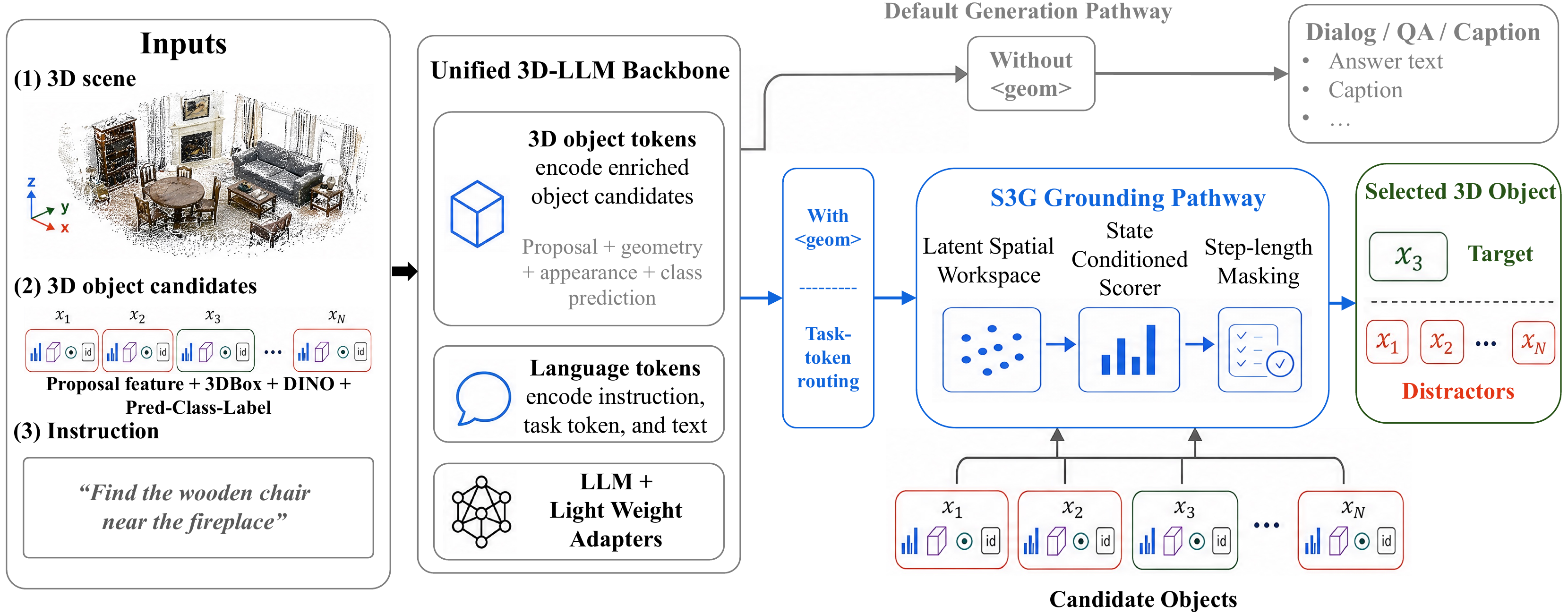}
  \caption{\textbf{SSR3D-LLM overview.}
  One 3D-LLM backbone handles language tasks and grounding.
  Without \texttt{\textless geom\textgreater}, it uses the default dialog/QA/caption route; with \texttt{\textless geom\textgreater}, it routes the instruction and Mask3D proposal representations to S3G for target selection and step-wise score traces.}
  \label{fig:ssr3dllm_overview}
  \vspace{-0.5em}
\end{figure*}

\begin{figure*}[t]
  \centering
  \includegraphics[width=\textwidth]{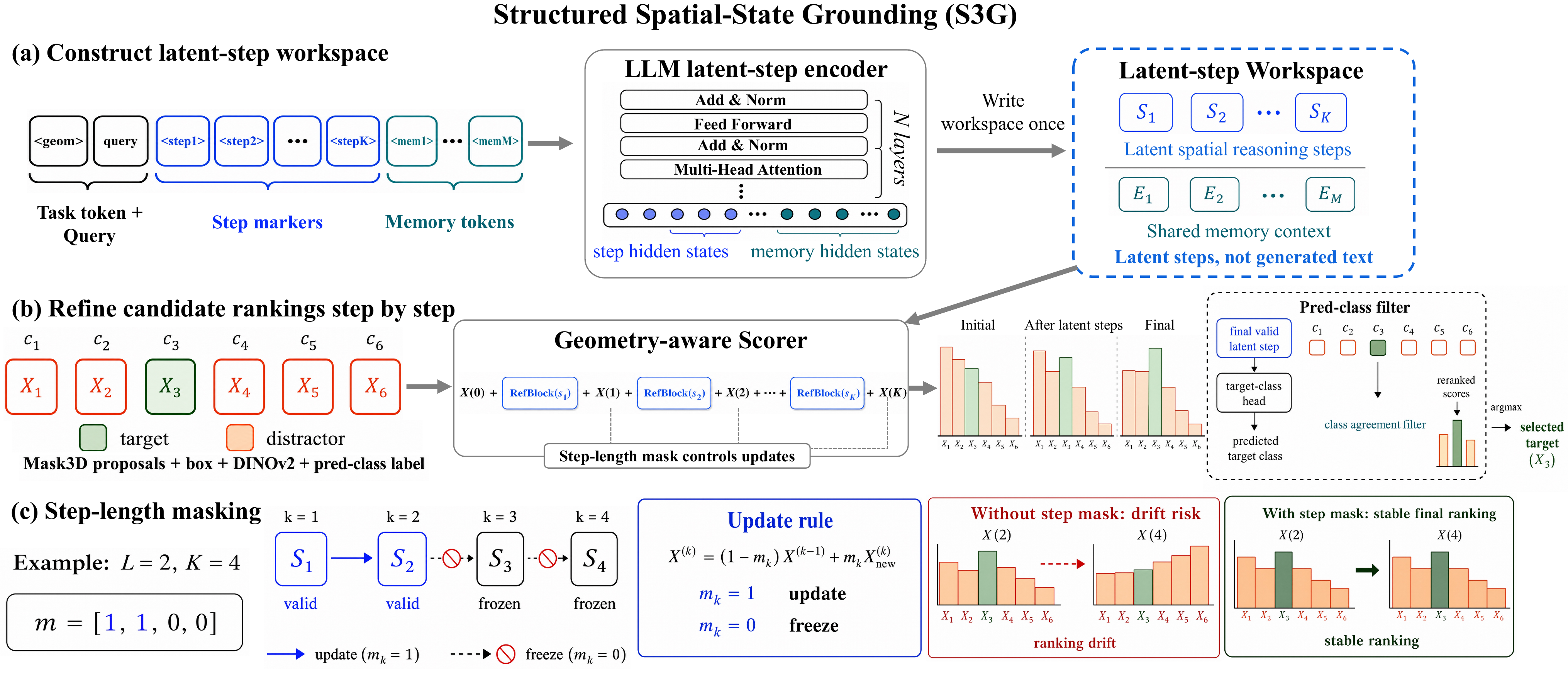}
  \caption{\textbf{S3G mechanism.}
  (a) The LLM writes a latent-step workspace in one forward pass by reading hidden states at reserved step markers and memory tokens.
  (b) A geometry-aware scorer reads these latent steps to refine Mask3D proposal rankings step by step, then a pred-class filter uses the predicted target class for final reranking.
  (c) Step-length masking keeps inactive steps from changing candidate states, supporting variable-length reasoning under a configurable step budget \(K\).}
  \label{fig:s3g_method_detail}
  \vspace{-0.5em}
\end{figure*}

SSR3D-LLM keeps the unified 3D-LLM backbone and the Mask3D-based candidate pipeline, but replaces the single \emph{QPG} pointer with a structured grounding route.
When \texttt{\textless geom\textgreater} is present, proposal, geometry, optional appearance, and proposal-side predicted-class evidence form object tokens; the LLM writes latent spatial reasoning steps and memory tokens; and a geometry-aware scorer refines candidate rankings step by step.
Without this task token, the model follows the default dialog/QA/caption path.
S3G is an interface-level design trained around Mask3D proposal tokens, latent-step supervision, and a candidate-level scoring head.
Figure~\ref{fig:ssr3dllm_overview} gives the route, while Figure~\ref{fig:s3g_method_detail} details the latent-step workspace, scorer, and mask.

\subsection{3D Candidate Space}
\label{sec:method_perception}
\vspace{-0.5em}

SSR3D-LLM grounds over point-cloud-derived Mask3D object proposals~\cite{Schult23ICRA}; the \emph{QPG} comparison uses the same proposal rows, proposal-level representation protocol, and evaluation mapping.
For each proposal \(i\), we combine its pooled proposal feature \(\mathbf{f}_i\), rotated-box geometry \(\mathbf{b}_i\), optional DINOv2 multi-view appearance \(\mathbf{a}_i\), and proposal-side predicted-class label \(\mathbf{c}_i\):
\begin{equation}
  \mathbf{x}^{(0)}_i = W_f \mathbf{f}_i + W_b \mathbf{b}_i + \alpha W_a \mathbf{a}_i + W_c \mathbf{c}_i,\quad
  \mathbf{X}^{(0)} = [\mathbf{x}^{(0)}_1;\ldots;\mathbf{x}^{(0)}_N].
\end{equation}
Here \(\mathbf{x}^{(0)}_i\in\mathbb{R}^{D}\) is the initial object candidate embedding for the \(i\)-th Mask3D proposal, \(\mathbf{X}^{(0)}\in\mathbb{R}^{N\times D}\) stacks the \(N\) candidate embeddings, \(W_f,W_b,W_a,W_c\) are learned projections, and $\alpha{=}0$ disables DINOv2.

\subsection{Constructing Latent Spatial Reasoning Steps}
\label{sec:method_llm}

The latent-step workspace uses \(K\) reserved \emph{step-marker} tokens and \(M\) learnable \emph{memory} tokens.
Step-marker hidden states serve as ordered latent spatial reasoning steps, while memory tokens carry shared query context.
For query \(u\), the inference-time grounding prompt is the token sequence \(p_{\text{pred}}=[\texttt{\textless geom\textgreater},u,\tau^{\mathrm{step}}_{1:K},\tau^{\mathrm{mem}}_{1:M}]\).
Let \(T_p\) be the number of prompt tokens before the memory tokens.
One LLM forward gives hidden states $\mathbf{H}\in\mathbb{R}^{(T_p+M)\times d_{\text{llm}}}$, which are projected into scorer space:
\begin{equation}
\begin{aligned}
\mathbf{s}_k &= \mathrm{LN}(W_s\,\mathbf{H}_{\mathrm{pos}(\tau^{\mathrm{step}}_k)}) \in \mathbb{R}^{D}, \quad k=1,\dots,K, \\
\mathbf{e}_j &= \mathrm{LN}(W_m\,\mathbf{H}_{T_p+j}) \in \mathbb{R}^{D}, \quad j=1,\dots,M,
\end{aligned}
\end{equation}
Here \(\mathrm{LN}\) is layer normalization, \(W_s,W_m\) are learned projections, and \(\mathrm{pos}(\cdot)\) returns a token position in \(p_{\text{pred}}\).
This yields latent step representations $\mathbf{S}\in\mathbb{R}^{K\times D}$ and memory representations $\mathbf{E}\in\mathbb{R}^{M\times D}$.

\subsection{Step-Wise Candidate Refinement}
\label{sec:method_listener}

Starting from \(\mathbf{X}^{(0)}\), the scorer maintains a candidate-state matrix \(\mathbf{X}^{(k)}=[\mathbf{x}^{(k)}_1;\ldots;\mathbf{x}^{(k)}_N]\in\mathbb{R}^{N\times D}\) after latent step \(k\).
At step $k$, it reads latent step representation $\mathbf{s}_k$ and memory $\mathbf{E}$, then applies a masked update with step mask \(m_k\in\{0,1\}\):
{\small
\begin{equation}
\label{eq:refer_update}
  \mathbf{X}^{(k)}_{\text{new}}=\mathrm{ReferBlock}_k\!\left(\mathbf{X}^{(k-1)},\mathbf{E},\mathbf{s}_k\right),\quad
  \mathbf{X}^{(k)}=(1-m_k)\mathbf{X}^{(k-1)}+m_k\mathbf{X}^{(k)}_{\text{new}}.
\end{equation}
}
ReferBlock lets candidates attend to memory under the current step cue, mix with each other, and undergo a small feed-forward refinement (Appendix Algorithm~\ref{alg:referblock}).
Because the initial candidate embeddings already combine Mask3D proposal features, rotated-box geometry, optional DINOv2 appearance, and predicted-class labels, the updates remain tied to the fixed proposal set.
Candidate logits are produced by a shared linear readout, \(\ell_i^{(k)}=\mathbf{w}_o^\top\mathbf{x}_i^{(k)}+b_o\), with final logits \(\boldsymbol{\ell}=\boldsymbol{\ell}^{(K)}\); intermediate logits form the trace.

\subsection{Handling Variable-Length Queries}
\label{sec:method_varlen}

Queries need different numbers of useful cues, so updates stop after the effective step length.
For effective length \(L\in\{1,\dots,K\}\), we set \(m_k{=}1\) for \(k\le L\) and \(m_k{=}0\) otherwise.
During training and diagnostic analysis, \(L\) can come from the referential-cue annotation.
For self-contained inference, a latent length predictor outputs \(q_k=\sigma(g_\psi(\mathbf{s}_k))\in(0,1)\), where \(\sigma\) is the sigmoid and \(g_\psi\) is a learned scalar head; \(q_k\) parameterizes whether step \(k\) is past the useful boundary.
With threshold \(\eta{=}0.5\), we set \(\hat L\) to the step before the first boundary hit, \(\hat L=\mathrm{clip}_{1:K}\!\left(\min(\{k-1:q_k\ge\eta\}\cup\{K\})\right)\), and build \(m_k=\mathbf{1}\{k\le\hat L\}\), where \(\mathrm{clip}_{1:K}\) clips to \(\{1,\dots,K\}\).
For example, \(q_1{=}0.05,q_2{=}0.12,q_3{=}0.83\) gives \(\hat L{=}2\), so updates after step 2 are frozen.
At inference, SSR3D-LLM extracts $(\mathbf{S},\mathbf{E})$ in one LLM forward, predicts this length mask, and applies Eq.~\eqref{eq:refer_update} to rank candidates.
The \emph{pred-class filter} uses the final valid step to predict a target class and reranks candidates whose proposal-side class agrees with it; the step logits $\boldsymbol{\ell}^{(k)}$ and mask values \(m_{1:K}\) form an inspectable non-textual trace.

\subsection{Training Objectives and Protocol}
\label{sec:method_training}

Training uses two supervision layers.
The answer supervision is the standard benchmark label \(i^\star\) or \(Y^\star\): the target candidate in ReferIt3D, and the target box or target set matched to Mask3D proposal rows for ScanRefer and Multi3DRef.
The referential cues are auxiliary latent-step supervision only.
For Sr3D, we deterministically convert the annotated target class, relation, and context-object fields into ordered cue phrases.
For Nr3D, which provides open-form utterances rather than structured context-object fields, an offline Qwen/vLLM annotator extracts a target, context-object, and relation cue sequence from the query and we normalize the object names to the training vocabulary when needed.
For ScanRefer and Multi3DRef, the same offline annotator uses the expression and, when available in the training annotation, the target class/name field to produce referential cue phrases in the same target/context-object/relation format.
These cues serve as training-only auxiliary supervision for the latent steps; inference uses the input query and fixed Mask3D proposals.
Concretely, if a training example yields \(L\) cue phrases, the first \(L\) latent steps are matched to their text embeddings and the remaining steps are masked out by \(m_k\).
At test time, the prompt contains only the original expression and reserved step/memory markers; the LLM must write the latent steps and predict the useful length from the query itself.

Training has three phases.
First, we warm-start the geometry-aware scorer with the candidate-ranking objective while using frozen LLM text features as language context.
Second, we train the LLM-side latent-step workspace so \(\mathbf{S}\) and \(\mathbf{E}\) replace the fixed text feature: latent step $k$ matches the $k$-th cue, and a global memory vector matches the full query.
Third, we continue with the inference-time prompt above, where the cue text is absent and the LLM must write the latent steps from the query alone.
A frozen BERT encoder provides step/global targets $\{\mathbf{t}_k\}$ and $\mathbf{t}_g$.
Only lightweight adapters and projections are updated by default.
For single-target examples, grounding logits use target cross-entropy:
\(
\mathcal{L}_{\text{ref}}=\mathrm{CE}\!\left(\mathrm{softmax}(\boldsymbol{\ell}), i^\star\right).
\)
For target-set examples, \(\mathcal{L}_{\text{ref}}\) is the corresponding multi-positive objective over matched candidates in \(Y^\star\), consistent with the set prediction metrics in Section~\ref{sec:prelim}.
Latent-step supervision uses cosine losses:
{\small
\begin{equation}
  \mathcal{L}_{\text{step}}=\frac{1}{\sum_k m_k}\sum_{k=1}^{K} m_k\big(1-\cos(\mathbf{s}_k,\mathbf{t}_k)\big),\quad
  \mathcal{L}_{\text{global}}=1-\cos(\mathbf{g},\mathbf{t}_g).
  \label{eq:latent_distill}
\end{equation}
} where $\mathbf{g}=\mathbf{e}_1$ is the global memory vector.
For ScanRefer-style identification, a target-class head on the last valid latent step adds $\mathcal{L}_{\text{cls}}$ and supplies the query-side class used by the pred-class filter at inference.
Here \(\mathrm{CE}\) is cross-entropy, \(\cos(\cdot,\cdot)\) is cosine similarity, and the \(\lambda\)'s are loss weights.
The total objective is
\(
\mathcal{L}=\mathcal{L}_{\text{ref}}+\lambda_{\text{step}}\mathcal{L}_{\text{step}}+\lambda_{\text{global}}\mathcal{L}_{\text{global}}+\lambda_{\text{cls}}\mathcal{L}_{\text{cls}}.
\)

\section{Experiments}
\label{sec:experiments}

\subsection{Setup and Protocol}
\label{sec:exp_benchmarks}
We evaluate on three main grounding benchmarks.
Nr3D/ReferIt3D (Table~\ref{tab:referit_main}) uses official candidate-set Top-1 accuracy and tests fine-grained object selection; ScanRefer and Multi3DRef (Table~\ref{tab:scan_multi_main}) test box localization under proposal noise and set-level grounding.
Additional comparisons, language examples, and trace diagnostics are in the appendix.

For ReferIt3D, we use the official supervised splits: 28,716 train / 7,485 test examples for Nr3D and 65,844 train / 17,726 test examples for Sr3D.
Across benchmarks, the supervised answer label is the dataset target object, box, or target set after proposal matching; auxiliary referential cues supervise only latent steps during training.
Unless stated otherwise, SSR3D-LLM uses \(K{=}4\) latent spatial reasoning steps and \(M{=}16\) memory tokens, with the Mask3D proposal representation from Section~\ref{sec:method_perception}.
At evaluation time, grounding uses the original expression and the Mask3D proposal set.
The latent steps and the pred-class filter are produced by the model from this input.
Language-capability experiments run the same checkpoint on ScanQA, Scan2Cap, object description, and embodied dialog through the default pathway without \texttt{\textless geom\textgreater}.

We organize baselines by system paradigm.
Specialized and task-specific grounding systems provide task-level references on the same benchmarks, showing how methods optimized directly for 3D grounding perform.
Reasoning-centric and agentic vision-LLM pipelines show what recent large-model grounding stacks can reach when they use rendered views, external prompting, or verification.
Our main interface comparison is the Unified 3D-LLMs block, where methods share the goal of preserving a general 3D-LLM while adding grounding.
Appendix Tables~\ref{tab:appendix_method_positioning}--\ref{tab:appendix_integrated_comparison} summarize the fuller interface context.
For unified baselines with internal object indices, the fixed mapping in Appendix~\ref{sec:appendix_build_up_index} converts predictions to the same ReferIt3D candidate protocol.

\subsection{Main Grounding Benchmark Results}
\label{sec:exp_results}
\label{sec:exp_main_referit}
Table~\ref{tab:referit_main} reports official Nr3D candidate-set accuracy.
Because Nr3D is not a native reported benchmark for Grounded 3D-LLM or Chat-Scene, the two unified-baseline rows are our ReferIt3D adaptations: both are finetuned on the official ReferIt3D training splits, and their internal pointer or object-ID outputs are converted to the official candidate-set prediction by the fixed mapping in Appendix~\ref{sec:appendix_build_up_index}.
Under this protocol, SSR3D-LLM reaches 50.3 overall accuracy.
The Grounded 3D-LLM comparison is the tighter interface test because it uses the \emph{QPG} readout under the same proposal-level representation protocol.
Chat-Scene provides a supplementary unified-model reference: it uses related proposal-level visual evidence, but differs in LLM backbone, object-token construction, grounding output interface, and training adaptation.

\begin{table*}[t]
    \centering
    \caption{\textbf{Nr3D grounding results.}
    We report official ReferIt3D candidate-set Top-1 accuracy.
    The Grounded 3D-LLM and Chat-Scene rows are ReferIt3D adaptations: both use grounding-specific finetuning on the official training split, and their internal pointer or object-ID outputs are evaluated through the same fixed instance-to-candidate mapping.
    Grounded 3D-LLM uses a \emph{QPG} interface under the same proposal-level representation protocol; Chat-Scene is included as a supplementary unified 3D-LLM reference with related proposal-level visual evidence but different backbone, object-token construction, output interface, and training adaptation.
    FT marks whether the method uses grounding-specific training or finetuning (\ftyes/\ftno).
    Backbone summaries provide context across methods with different perception front ends and visual evidence.
    ``--'' means the original paper does not report that Nr3D split; bold numbers mark the best result within the Unified 3D-LLMs.}
    \label{tab:referit_main}
    \vspace{0.15em}
    \begingroup
    \scriptsize
    \setlength{\tabcolsep}{1.0pt}
    \renewcommand{\arraystretch}{1.06}
    \resizebox{\textwidth}{!}{%
    \begin{tabular}{@{}P{3.85cm}P{2.05cm}c|*{5}{>{\centering\arraybackslash}p{0.86cm}}@{}}
        \toprule
        \textbf{Method} & \textbf{Backbone} & \textbf{FT} & \textbf{All} & \textbf{Easy} & \textbf{Hard} & \textbf{V-Dep.} & \textbf{V-Ind.} \\
        \midrule
        \rowcolor{referitsectionbg}\multicolumn{8}{@{}l@{}}{\textit{Specialized / task-specific grounding systems}}\\
        ReferIt3DNet (ECCV'20) & 3D listener & \ftyes & 35.6 & 43.6 & 27.9 & 32.5 & 37.1 \\
        WS-3DVG (ICCV'23) & 3DVG net & \ftyes & 22.5 & 27.3 & 18.0 & 21.9 & 22.9 \\
        M3DRef-CLIP (ICCV'23) & CLIP & \ftyes & 49.4 & 55.6 & 43.4 & 42.3 & 52.9 \\
        ZSVG3D (CVPR'24) & GPT-4 + CLIP & \ftno & 39.0 & 46.5 & 31.7 & 36.8 & 40.0 \\
        MiKASA (CVPR'24) & Transformer & \ftyes & 64.4 & 69.7 & 59.4 & 65.4 & 64.0 \\
        SeeGround (CVPR'25) & Qwen2-VL-72B & \ftno & 46.1 & 54.5 & 38.3 & 42.3 & 48.2 \\
        LaSP (EMNLP'25) & LLM program & \ftno & 52.9 & 60.7 & 45.3 & 49.2 & 54.7 \\
        ViGOR (WACV'25) & 3D listener & \ftyes & 59.7 & 66.6 & 53.1 & 59.2 & 59.9 \\
        SORT3D (arXiv'25) & LLM toolbox & \ftno & 62.0 & -- & -- & 56.6 & 64.3 \\
        \midrule
        \rowcolor{referitsectionbg}\multicolumn{8}{@{}l@{}}{\textit{Reasoning-centric / agentic large-model pipelines}}\\
        VLM-Grounder (CVPR'24) & GPT-4V & \ftno & 48.0 & 55.2 & 39.5 & 45.8 & 49.4 \\
        SPAZER (arXiv'25) & GPT-4o & \ftno & 63.8 & 68.0 & 58.8 & 59.9 & 66.2 \\
        \midrule
        \rowcolor{referitsectionbg}\multicolumn{8}{@{}l@{}}{\textit{Unified 3D-LLMs}}\\
        Chat-Scene (NeurIPS'24) & Vicuna-7B & \ftyes & 25.9 & 31.0 & 21.4 & 21.6 & 28.3 \\
        Grounded 3D-LLM (arXiv'24) & Tiny-Vicuna-1B & \ftyes & 32.8 & 40.3 & 25.5 & 29.7 & 34.3 \\
        \rowcolor{referitreferencebg}
        \textbf{SSR3D-LLM} & \textbf{Tiny-Vicuna-1B} & \ftyes & \textbf{50.3} & \textbf{61.0} & \textbf{39.9} & \textbf{46.5} & \textbf{52.1} \\
        \bottomrule
    \end{tabular}
    }%
    \endgroup
\end{table*}

\begin{table*}[t]
    \centering
    \caption{\textbf{ScanRefer and Multi3DRef evaluation.}
    ScanRefer reports Acc@0.25IoU / Acc@0.50IoU; Multi3DRef reports F1@0.25IoU / F1@0.50IoU.
    These benchmarks evaluate 3D box localization and set-level grounding under proposal noise.
    FT marks whether the method uses grounding-specific training or finetuning (\ftyes/\ftno).
    Backbone summaries provide context across methods with different visual evidence and reasoning stacks.
    ``--'' means the original paper does not report that benchmark/metric, or the method was not evaluated for the corresponding output type (single-object boxes vs. target sets); bold numbers mark the best result within the Unified 3D-LLMs.}
    \label{tab:scan_multi_main}
    \vspace{0.15em}
    \begingroup
    \small
    \setlength{\tabcolsep}{0pt}
    \renewcommand{\arraystretch}{1.05}
    \begin{tabular*}{\textwidth}{@{}l@{\extracolsep{\fill}}l@{\extracolsep{0pt}\hspace{0.35em}}c@{\hspace{0.22em}}|@{\hspace{0.18em}}*{2}{>{\centering\arraybackslash}p{1.24cm}@{\hspace{0.24em}}}|@{\hspace{0.18em}}*{2}{>{\centering\arraybackslash}p{1.24cm}@{\hspace{0.24em}}}@{}}
        \toprule
        \multirow{2}{*}{\textbf{Method}} &
        \multirow{2}{*}{\textbf{Backbone}} &
        \multirow{2}{*}{\textbf{FT}} &
        \multicolumn{2}{c|}{\textbf{ScanRefer}} &
        \multicolumn{2}{c}{\textbf{Multi3DRef}} \\
        \cmidrule(lr){4-5}\cmidrule(l){6-7}
        & & & \textbf{A@0.25} & \textbf{A@0.50} & \textbf{F1@0.25} & \textbf{F1@0.50} \\
        \midrule
        \rowcolor{referitsectionbg}\multicolumn{7}{@{}l@{}}{\textit{Specialized / task-specific grounding systems}}\\
        WS-3DVG (ICCV'23) & 3DVG net & \ftyes & 27.37 & 21.96 & -- & -- \\
        M3DRef-CLIP (ICCV'23) & CLIP & \ftyes & -- & -- & 42.8 & 38.4 \\
        ZSVG3D (CVPR'24) & GPT-4 + CLIP & \ftno & 36.4 & 32.7 & -- & -- \\
        SeeGround (CVPR'25) & Qwen2-VL-72B & \ftno & 44.1 & 39.4 & -- & -- \\
        \midrule
        \rowcolor{referitsectionbg}\multicolumn{7}{@{}l@{}}{\textit{Reasoning-centric / agentic large-model pipelines}}\\
        VLM-Grounder (CVPR'24) & GPT-4V & \ftno & 51.6 & 32.8 & -- & -- \\
        SPAZER (arXiv'25) & GPT-4o & \ftno & 57.2 & 48.8 & -- & -- \\
        GPT4Scene (arXiv'25) & GPT-4o & \ftno & 40.5 & 36.7 & 45.4 & 42.1 \\
        GPT4Scene-HD (arXiv'25) & GPT-4o & \ftno & 50.9 & 46.4 & 53.7 & 50.0 \\
        GPT4Scene-HDM (arXiv'25) & GPT-4o & \ftno & 62.6 & 57.0 & 64.5 & 59.8 \\
        \midrule
        \rowcolor{referitsectionbg}\multicolumn{7}{@{}l@{}}{\textit{Unified 3D-LLMs}}\\
        Chat-Scene (NeurIPS'24) & Vicuna-7B & \ftyes & 55.5 & 50.2 & 57.1 & 52.4 \\
        Grounded 3D-LLM (arXiv'24) & Tiny-Vicuna-1B & \ftyes & 48.6 & 44.0 & 44.7 & 40.8 \\
        \rowcolor{referitreferencebg}
        \textbf{SSR3D-LLM} & Tiny-Vicuna-1B & \ftyes & \textbf{58.7} & \textbf{53.9} & \textbf{63.1} & \textbf{57.9} \\
        \bottomrule
    \end{tabular*}
    \endgroup
\end{table*}

Table~\ref{tab:scan_multi_main} extends the evaluation to ScanRefer and Multi3DRef.
On ScanRefer, SSR3D-LLM reaches 58.7 / 53.9 Acc@0.25 / Acc@0.50, above Chat-Scene's 55.5 / 50.2 and Grounded 3D-LLM's 48.6 / 44.0 within the unified-model block.
On Multi3DRef, it reaches 63.1 / 57.9 F1@0.25 / F1@0.50, again improving over Chat-Scene (57.1 / 52.4) and Grounded 3D-LLM (44.7 / 40.8).
Thus, SSR3D-LLM improves grounding across box localization and set-level reference within the unified 3D-LLM family, while the higher GPT4Scene-HDM numbers indicate that stronger rendered-view reasoning stacks remain complementary to our 3D-native interface design.

On language-capability experiments, the same checkpoint without \texttt{\textless geom\textgreater} remains comparable to the Grounded 3D-LLM \emph{QPG} baseline (Dialog 1.090 vs.\ 1.247 CIDEr, QA 19.83 vs.\ 20.11 EM, Cap 0.658 vs.\ 0.626 CIDEr, Desc 1.027 vs.\ 1.052 CIDEr).
The gains therefore do not come from turning the model into a grounding-only listener; examples are in Appendix Section~\ref{sec:appendix_cap_examples}.

\subsection{Analysis and Ablations}
\label{sec:exp_analysis}
Table~\ref{tab:compact_diagnostics} evaluates the main S3G step-wise mechanisms while holding the proposal representation fixed.
Removing the step mask drops Nr3D from 50.3 to 46.9 and Sr3D from 67.5 to 59.1.
The learned length predictor is close on Sr3D (67.0 vs.\ 67.5) but weaker on Nr3D (47.2 vs.\ 50.3), while shuffling or reversing valid latent steps sharply degrades performance.
These diagnostics show that the scorer uses ordered latent steps and that padded steps must be kept inert.

\begin{figure}[!t]
    \centering
    \makebox[\textwidth][c]{\includegraphics[width=1.04\textwidth]{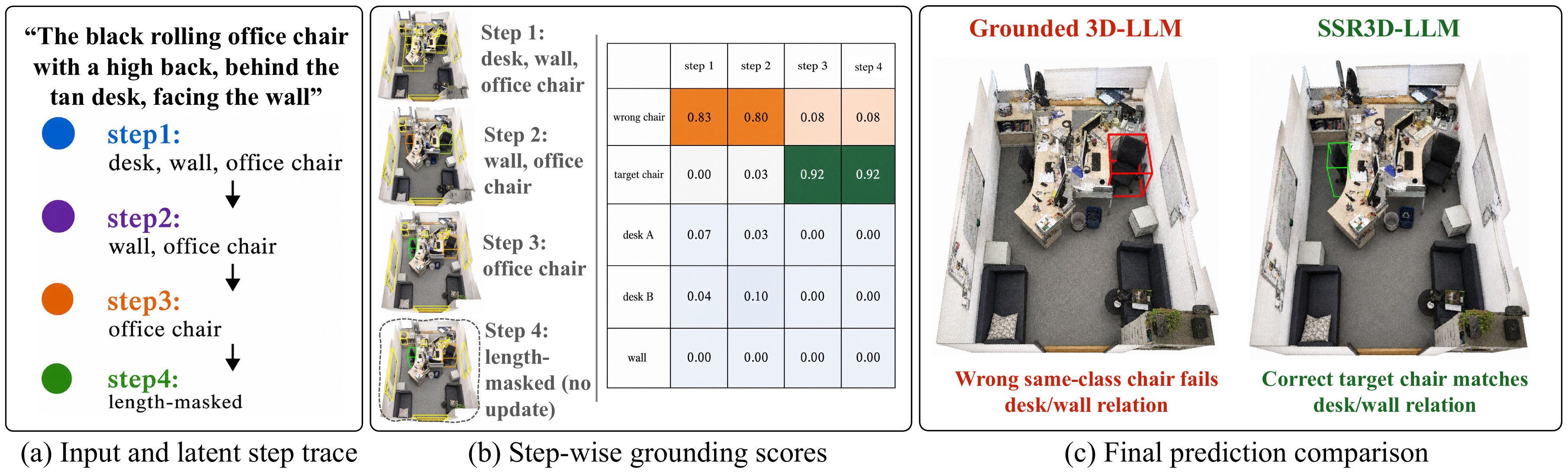}}
    \caption{\textbf{Qualitative grounding case.}
    A real ScanRefer example shows the query, step-wise candidate scores, proposal boxes, and baseline vs.\ SSR3D-LLM predictions.
    The masked fourth slot preserves the step-3 ranking; cue labels are visualization-only annotations for the figure.}
    \label{fig:qual_trace_card}
    \vspace{-1.0em}
\end{figure}

Figure~\ref{fig:qual_trace_card} illustrates the same mechanism on a concrete case.
The single-pointer baseline can lock onto a visually plausible but relationally wrong object, while S3G first stabilizes context-object evidence, then uses the later spatial cue to separate competing candidates.
The visualization is meant to show what the score trace adds beyond the final prediction: each step exposes which candidates remain plausible and when the target becomes separated.

\begin{table}[t]
    \centering
    \caption{\textbf{Ablations on SSR3D-LLM.}
    Rows show changes from Full under the same proposal representation; \(\Delta\) columns report absolute drops or gains.}
    \label{tab:compact_diagnostics}
    \vspace{0.05em}
    \begingroup
        \small
        \setlength{\tabcolsep}{3.4pt}
        \renewcommand{\arraystretch}{1.0}
        \begin{tabular*}{\textwidth}{@{}P{2.25cm}P{5.75cm}@{\hspace{0.8em}}r@{\extracolsep{\fill}}rrr@{}}
            \toprule
            \textbf{Variant} & \textbf{Change from Full} & \textbf{Nr3D} & \textbf{\(\Delta\)Nr3D} & \textbf{Sr3D} & \textbf{\(\Delta\)Sr3D} \\
            \midrule
            \rowcolor{referitreferencebg}
            Full & Reference S3G setting & \textbf{50.3} & -- & \textbf{67.5} & -- \\
            \(+\) pred. length & Predict useful length with a length gate & 47.2 & $-3.1$ & 67.0 & $-0.5$ \\
            \(-\) step mask & Remove step mask; use fixed-\(K\) updates & 46.9 & $-3.4$ & 59.1 & $-8.4$ \\
            \(+\) shuffle order & Shuffle valid latent steps & 46.2 & $-4.1$ & 56.8 & $-10.7$ \\
            \(+\) reverse order & Reverse valid latent steps & 41.7 & $-8.6$ & 46.3 & $-21.2$ \\
            \bottomrule
        \end{tabular*}
    \endgroup
    \vspace{-0.8em}
\end{table}

\begin{table}[t]
    \centering
    \caption{\textbf{Grounding readout runtime.}
    All rows use precomputed proposal-level inputs and measure only the grounding readout from object/query representations to the final prediction.
    Timing excludes Mask3D proposal extraction, DINOv2 feature generation, query-side preprocessing, and data loading.
    Chat-Scene is included as a broader unified-3D-LLM reference rather than a same-protocol runtime baseline; Appendix Section~\ref{sec:appendix_repro_assets_responsible} details the measurement protocol.}
    \label{tab:runtime_matched}
    \vspace{0.2em}
    \begingroup
    \small
    \setlength{\tabcolsep}{3.2pt}
    \renewcommand{\arraystretch}{1.05}
    \begin{tabular*}{\textwidth}{@{}P{1.75cm}P{3.05cm}@{\extracolsep{\fill}}ccc@{}}
        \toprule
        \textbf{Dataset} & \textbf{Method} & \textbf{Latency (s/query)} & \textbf{Throughput (query/s)} & \textbf{Peak Mem. (GB)} \\
        \midrule
        \multirow{3}{*}{Nr3D}
        & Grounded 3D-LLM & 1.1610 & 0.861 & 3.159 \\
        & Chat-Scene & 1.0787 & 0.927 & 19.741 \\
        & \textbf{SSR3D-LLM} & \textbf{0.0363} & \textbf{27.585} & 5.792 \\
        \midrule
        \multirow{3}{*}{ScanRefer}
        & Grounded 3D-LLM & 0.9882 & 1.012 & 3.192 \\
        & Chat-Scene & 1.4412 & 0.694 & 19.740 \\
        & \textbf{SSR3D-LLM} & \textbf{0.0388} & \textbf{25.765} & 5.795 \\
        \midrule
        \multirow{3}{*}{Multi3DRef}
        & Grounded 3D-LLM & 1.0166 & 0.984 & 3.186 \\
        & Chat-Scene & 1.2532 & 0.798 & 19.674 \\
        & \textbf{SSR3D-LLM} & \textbf{0.0380} & \textbf{26.299} & 5.792 \\
        \bottomrule
    \end{tabular*}
    \endgroup
    \vspace{-0.7em}
\end{table}

Table~\ref{tab:runtime_matched} isolates the cost of the grounding readout after proposal-level representations have been prepared.
The measurement covers readout-stage latency rather than end-to-end system latency: proposal extraction, DINOv2 feature generation, query-side preprocessing, and data loading are excluded.
Under this readout-only protocol, SSR3D-LLM is substantially faster than the single-pointer \emph{QPG} readout and the Chat-Scene reference across all three benchmarks.
The result supports a narrow efficiency claim: latent spatial reasoning steps do not require autoregressive trace decoding or repeated LLM calls, and the final candidate ranking can be computed as a lightweight one-pass readout.

Additional runtime diagnostics, GT-rank flow, score heatmaps, and reliability plots are in Appendix Table~\ref{tab:chainlen_efficiency} and Figures~\ref{fig:gtrankflow}--\ref{fig:macro_error_modes_full}.

\noindent\textbf{Remark.}
The experiments isolate S3G over fixed Mask3D candidates and proposal-level representations; stronger proposal sources, rendered views, and richer visual inputs remain complementary.

\FloatBarrier
\vspace{-0.5em}
\section{Conclusion}
\label{sec:conclusion}

We propose SSR3D-LLM, which reframes fine-grained grounding in unified 3D-LLMs as an interface problem.S3G replaces single-pointer \emph{QPG} with latent spatial reasoning steps: the LLM writes a workspace from the query, and a geometry-aware scorer ranks fixed Mask3D candidates step by step.
Auxiliary referential cues supervise this workspace during training, but inference uses only the original query and object proposals.
Across ReferIt3D, ScanRefer, and Multi3DRef, this route improves over \emph{QPG} and prior unified 3D-LLM baselines while preserving the language-task pathway.
The current system still depends on upstream proposal quality and uses a fixed latent-step budget; extending S3G to stronger proposal sources, richer visual inputs, adaptive or longer latent steps, and broader 3D-LLM backbones are natural next steps.

\FloatBarrier

\balance
\bibliographystyle{plainnat}
\bibliography{ref}

\clearpage
\appendix
\onecolumn
\section{Additional Results and Ablations}
\label{sec:appendix}

\subsection{Discussion, Limitations, and Future Work}
\label{sec:appendix_discussion_limitations}

\paragraph{Scope of the claim.}
SSR3D-LLM targets the grounding interface inside unified, instance-centric 3D-LLMs.
The main comparison therefore emphasizes unified 3D-LLM baselines that expose grounding through compact object-id or pointer-style readouts.
Specialized grounding systems, agentic pipelines, and stronger vision-LLM stacks remain important broader references, but they often use different visual front ends, rendered views, external verification loops, or supervision protocols.

\paragraph{Limitations.}
Our experiments use fixed Mask3D proposal sets and ScanNet-style indoor benchmarks, so grounding quality is still bounded by upstream proposal recall, instance segmentation quality, and the domain coverage of the underlying datasets.
Rotated-box geometry, DINOv2 appearance, and proposal-side predicted classes improve the proposal representation, but they do not replace the proposal generator.
The latent steps are also shaped by training-time referential cues, whose coverage and noise can affect how cleanly the step-wise trace aligns with human-interpretable context objects and relations.
We use a fixed maximum step budget \(K{=}4\) with step-length masking; this covers the evaluated benchmarks but may limit expressions with longer or more nested relational chains, motivating adaptive or larger step budgets in future work.

\paragraph{Future work.}
Promising directions include combining S3G with stronger open-vocabulary 3D perception, extending evaluation beyond static indoor ScanNet-style scenes, and studying closed-loop use in embodied agents where grounding errors can affect downstream actions.
Another direction is to reduce reliance on offline referential-cue construction by learning latent-step structure from weaker or self-supervised signals while preserving the default language pathway of unified 3D-LLMs.

\subsection{Additional Qualitative Visualizations}
\label{sec:appendix_additional_visualizations}

\begin{figure*}[p]
  \centering
  \includegraphics[width=\textwidth]{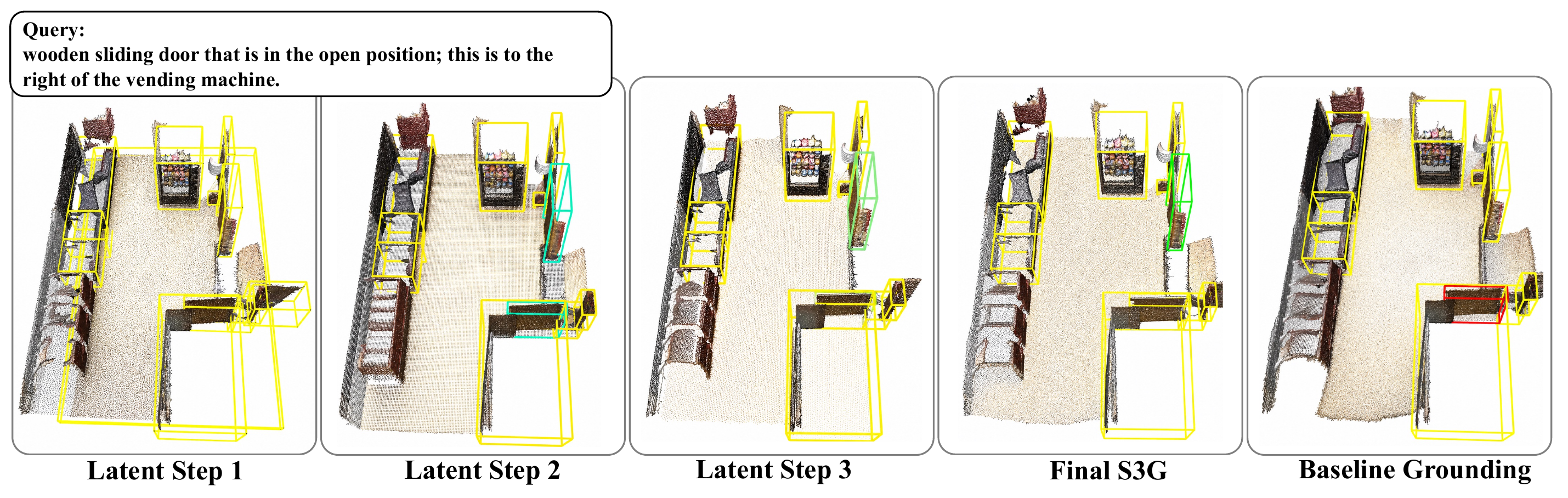}
  \caption{\textbf{Additional qualitative visualization I.}}
  \label{fig:appendix_visualization_1}
\end{figure*}

\begin{figure*}[p]
  \centering
  \includegraphics[width=\textwidth]{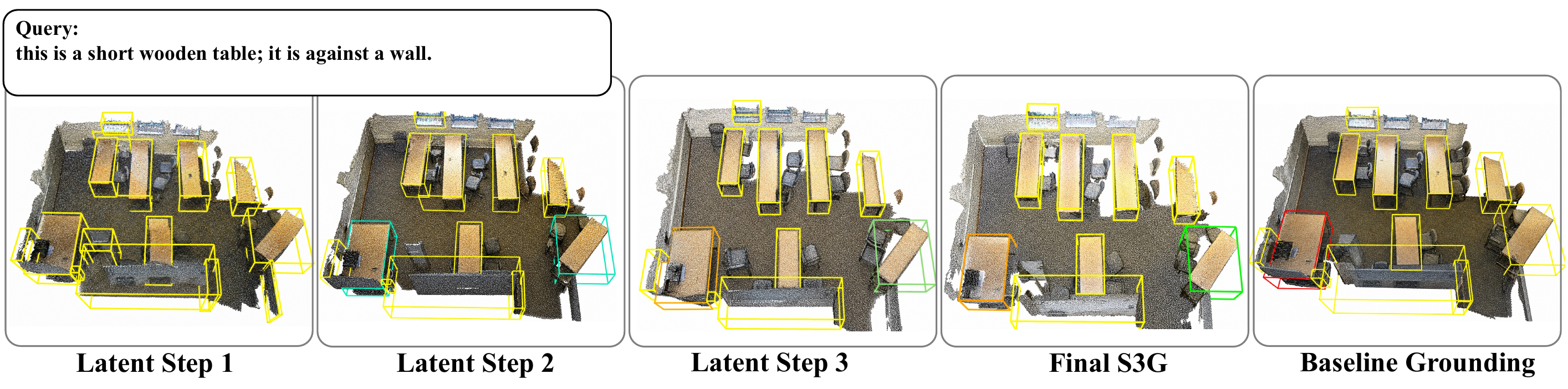}
  \caption{\textbf{Additional qualitative visualization II.}}
  \label{fig:appendix_visualization_2}
\end{figure*}

\begin{figure*}[p]
  \centering
  \includegraphics[width=\textwidth]{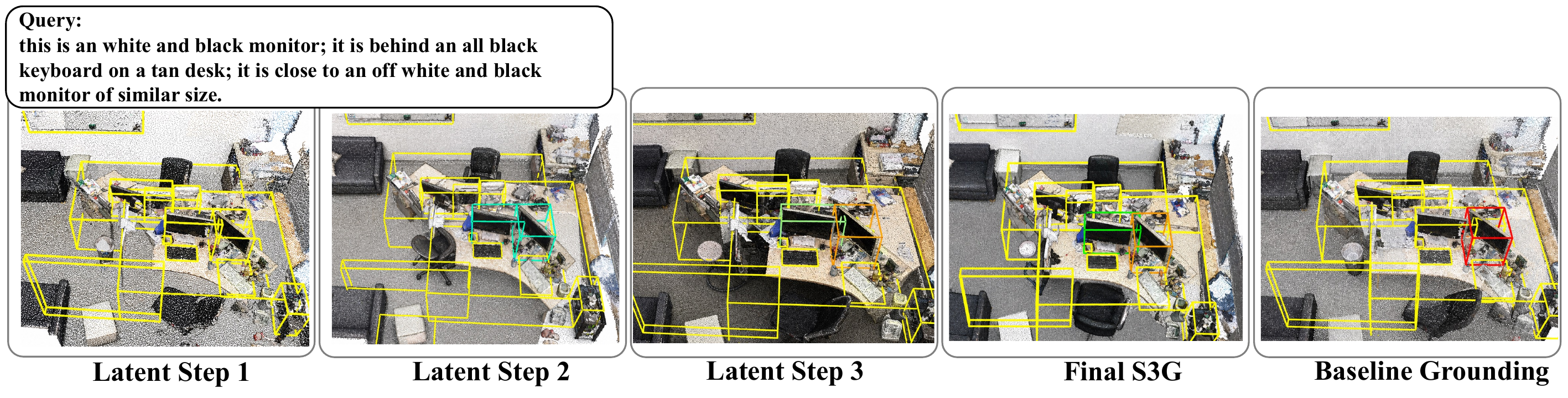}
  \caption{\textbf{Additional qualitative visualization III.}}
  \label{fig:appendix_visualization_3}
\end{figure*}

\begin{figure*}[p]
  \centering
  \includegraphics[width=\textwidth]{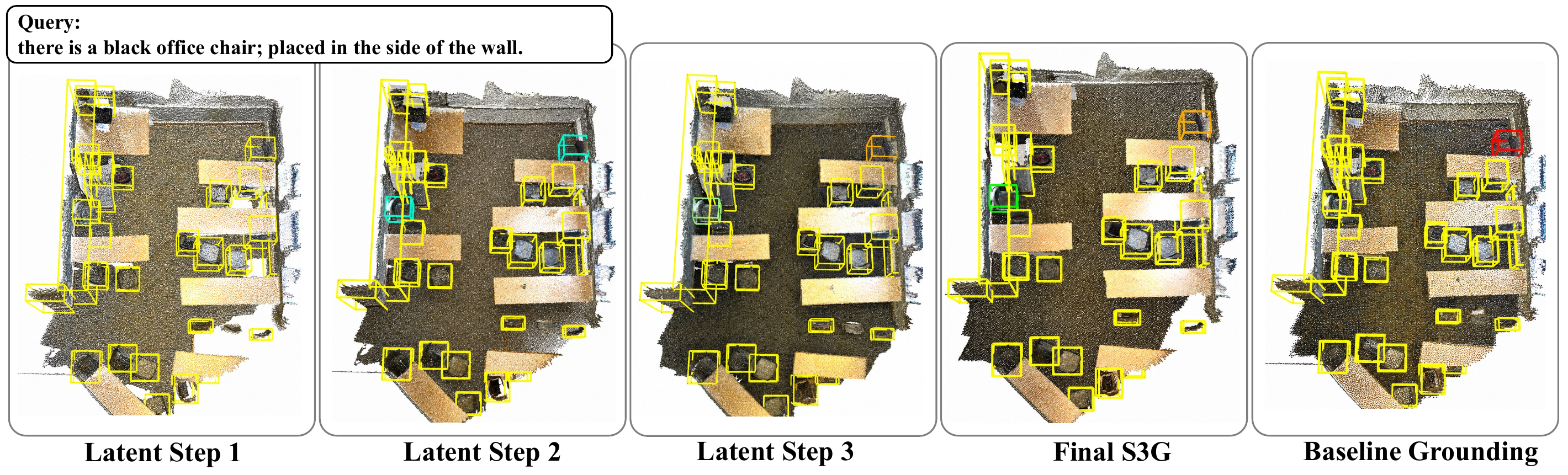}
  \caption{\textbf{Additional qualitative visualization IV.}}
  \label{fig:appendix_visualization_4}
\end{figure*}

Figures~\ref{fig:appendix_visualization_1}--\ref{fig:appendix_visualization_4} provide additional qualitative grounding cases.
They complement the main qualitative example by showing how candidate rankings evolve across latent steps and how the final prediction is selected from the fixed Mask3D proposal set.
In all four examples, the final S3G selection matches the ground-truth target object.
Across these visualizations, yellow 3D boxes denote the Mask3D proposal candidates considered by the grounding model.
Cyan and orange boxes highlight intermediate proposals that receive high scores or remain competitive at a latent step, green boxes mark the final S3G selection, and red boxes mark the baseline prediction when it selects a different object.
The colors are visualization overlays for readability; the model operates on the same fixed proposal representations described in the 3D candidate-space subsection.

\subsection{ReferBlock Details}
\label{sec:appendix_referblock}

\paragraph{What it does.}
At each step $k$, $\mathrm{ReferBlock}_k(\cdot)$ takes the current candidate tokens $\mathbf{X}^{(k-1)}$ and updates them using (i) the step cue $\mathbf{s}_k$ and (ii) the shared memory $\mathbf{E}$.
Intuitively, candidates first ``read'' the global instruction from $\mathbf{E}$, then mix information across candidates, and finally apply a small feed-forward refinement.
In Algorithm~\ref{alg:referblock}, \(\mathrm{LN}\) denotes layer normalization, \(\mathrm{CrossAttn}\) and \(\mathrm{SelfAttn}\) are standard multi-head attention blocks, and \(\mathrm{MLP}\) is a feed-forward block.

\begin{algorithm}[t]
  \caption{\textbf{ReferBlock} computation at step $k$.}
  \label{alg:referblock}
  \begin{algorithmic}[1]
    \REQUIRE Candidate tokens $\mathbf{X}^{(k-1)}\in\mathbb{R}^{N\times D}$, memory $\mathbf{E}\in\mathbb{R}^{M\times D}$, step cue $\mathbf{s}_k\in\mathbb{R}^{D}$
    \ENSURE Updated tokens $\mathbf{X}^{(k)}_{\text{new}}\in\mathbb{R}^{N\times D}$
    \STATE $\mathbf{Q}\leftarrow \mathbf{X}^{(k-1)}$
    \STATE $\mathbf{Q}\leftarrow \mathbf{Q} + \mathrm{CrossAttn}(\mathrm{LN}(\mathbf{Q}),\, \mathrm{LN}(\mathbf{E}),\, \mathbf{s}_k)$ \COMMENT{read instruction with step cue}
    \STATE $\mathbf{Q}\leftarrow \mathbf{Q} + \mathrm{SelfAttn}(\mathrm{LN}(\mathbf{Q}))$ \COMMENT{mix candidates}
    \STATE $\mathbf{Q}\leftarrow \mathbf{Q} + \mathrm{MLP}(\mathrm{LN}(\mathbf{Q}))$
    \STATE $\mathbf{X}^{(k)}_{\text{new}}\leftarrow \mathbf{Q}$
  \end{algorithmic}
\end{algorithm}

\subsection{Unified Baseline Adaptation and Index-to-Candidate Evaluation}
\label{sec:appendix_build_up_index}
Grounded 3D-LLM and Chat-Scene use internal object-index outputs, so Table~\ref{tab:referit_main} evaluates ReferIt3D adaptations under the official candidate-set protocol.
Both adapted unified baselines are finetuned on the official ReferIt3D training splits and evaluated on the official test split.
Their grounding heads produce internal identifiers--a \emph{QPG} backbone index for Grounded 3D-LLM, and an object-ID style output for Chat-Scene--rather than a native score for every ReferIt3D candidate.
To evaluate these outputs under the official candidate-set protocol, we map the predicted internal identifier to the corresponding candidate object using a fixed mapping derived from the same Mask3D perception outputs.

\paragraph{Definition.}
We define a \emph{reference index} \(j^\star\) by mapping the ReferIt3D target instance into the backbone index space using a fixed instance-to-index mapping built from the same Mask3D perception outputs.
Given the model-predicted top-1 index \(\hat{j}\), the prediction is correct iff \(\hat{j}=j^\star\); invalid indices count as misses.

\paragraph{Evaluation protocol.}
All evaluations use (i) the official ReferIt3D splits and the same context construction/sampling procedure, (ii) the same fixed mapping (derived from the same Mask3D perception outputs) to define \(j^\star\), and (iii) the same evaluation pipeline (mapping, filtering, and hit criterion), where we only swap the evaluated model.
For the Grounded 3D-LLM \emph{QPG} comparison, the candidate representation also uses the same CLASP-pretrained Mask3D features, rotated-box geometry, and DINOv2 appearance used by SSR3D-LLM; only the grounding readout changes.
The Chat-Scene row uses related proposal-level visual evidence but keeps the corresponding Chat-Scene object-token pipeline, backbone, output interface, and training adaptation, so it is a supplementary unified-model reference rather than the matched interface baseline.

\subsection{Auxiliary Representation Visualization}
\label{sec:appendix_tsne}

The main evidence comes from behavioral comparisons under the same candidate pipeline, proposal-level representation protocol, and backbone.
The following visualization provides an auxiliary qualitative view of the learned pointer representation.

\begin{figure}[t]
  \centering
  \includegraphics[width=\textwidth]{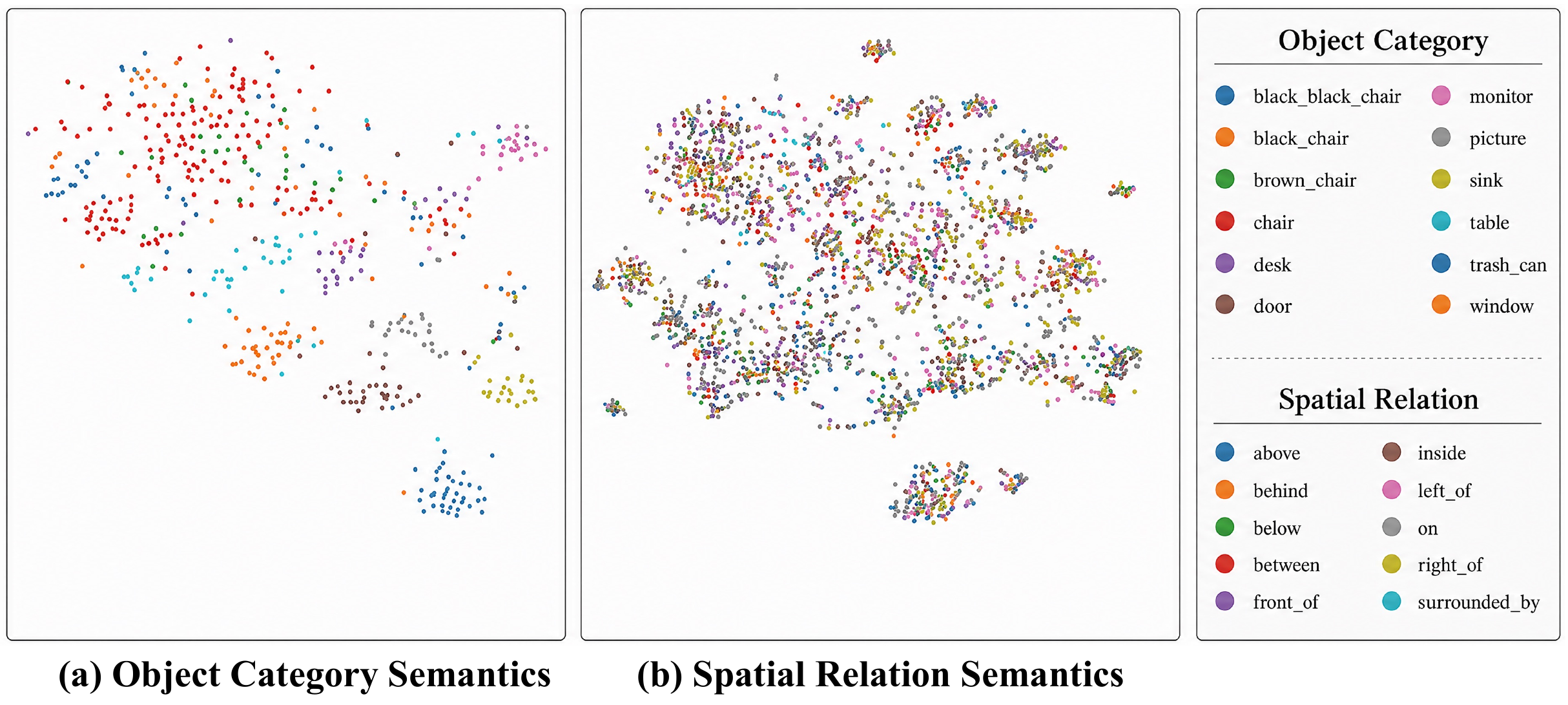}
  \caption{\textbf{Auxiliary representation visualization for \emph{QPG}.}
  The \emph{QPG} representation separates object-category semantics more clearly than spatial-relation semantics.
  We use this as a qualitative view of the single-step bottleneck; Section~\ref{sec:exp_main_referit} evaluates its effect on object selection.}
  \label{fig:tsne_ref_sem_spatial}
  \vspace{-0.6em}
\end{figure}

\subsection{Additional Comparisons and Transfer Evidence}
\label{sec:appendix_positioning}

\paragraph{What these tables add.}
The main paper keeps the core evidence in two result tables and one compact ablation table.
This appendix subsection expands the experimental picture along four axes: interface positioning across method families, the Sr3D split-level breakdown, cross-backbone transfer on Chat-Scene, and the cost of producing latent-step traces.
Together, these tables support the same central conclusion: S3G changes the grounding readout inside unified 3D-LLMs, while broader specialized or agentic systems provide useful literature context under different visual and inference protocols.

\paragraph{Interface positioning.}
Tables~\ref{tab:appendix_method_positioning} and~\ref{tab:appendix_integrated_comparison} separate task-specific grounding systems, reasoning-centric pipelines, and unified 3D-LLMs.
The key distinction is that SSR3D-LLM keeps the unified dialog/QA/captioning setting, but replaces compact single-readout grounding with latent-step candidate scoring over fixed Mask3D proposals.

\begin{table*}[t]
    \centering
    \caption{\textbf{Grounding interfaces across method families.}
    Comparison of task coverage, whether grounding uses an intermediate reasoning workspace, and what representation the final grounding readout uses.}
    \label{tab:appendix_method_positioning}
    \vspace{0.2em}
    \begingroup
    \small
    \renewcommand{\arraystretch}{1.05}
    \begin{tabularx}{\textwidth}{@{}P{2.35cm}*{4}{>{\centering\arraybackslash}p{0.76cm}}P{2.55cm}>{\raggedright\arraybackslash}X@{}}
        \toprule
        \textbf{Method} & \textbf{3DVG} & \textbf{QA} & \textbf{Cap.} & \textbf{Dialog} & \textbf{Reasoning workspace} & \textbf{Grounding readout} \\
        \midrule
        \rowcolor{referitsectionbg}\multicolumn{7}{@{}l@{}}{\textit{Specialized / task-specific grounding systems}}\\
        ZSVG3D & $\checkmark$ & $\times$ & $\times$ & $\times$ & None & Feature / text matching \\
        SeeGround & $\checkmark$ & $\times$ & $\times$ & $\times$ & None & VLM candidate matching \\
        LaSP & $\checkmark$ & $\times$ & $\times$ & $\times$ & None & Spatial feature matching \\
        SORT3D & $\checkmark$ & $\times$ & $\times$ & $\times$ & None & Object-relation matching \\
        \rowcolor{referitsectionbg}\multicolumn{7}{@{}l@{}}{\textit{Reasoning-centric / agentic large-model pipelines}}\\
        VLM-Grounder & $\checkmark$ & $\times$ & $\times$ & $\times$ & External prompts & 2D-to-3D verified matching \\
        SPAZER & $\checkmark$ & $\times$ & $\times$ & $\times$ & External agent & Agentic candidate verification \\
        GPT4Scene & $\checkmark$ & $\times$ & $\times$ & $\times$ & External prompts & Scene-level verification \\
        \rowcolor{referitsectionbg}\multicolumn{7}{@{}l@{}}{\textit{Unified 3D-LLMs}}\\
        Grounded 3D-LLM & $\checkmark$ & $\checkmark$ & $\checkmark$ & $\checkmark$ & None & Single-token \emph{QPG} pointer \\
        Chat-Scene & $\checkmark$ & $\checkmark$ & $\checkmark$ & $\checkmark$ & None & Single-step object-ID token \\
        \rowcolor{referitreferencebg}
        \textbf{SSR3D-LLM (ours)} & $\boldsymbol{\checkmark}$ & $\boldsymbol{\checkmark}$ & $\boldsymbol{\checkmark}$ & $\boldsymbol{\checkmark}$ & \textbf{Latent steps} & \textbf{Reasoning-state scoring} \\
        \bottomrule
    \end{tabularx}
    \endgroup
\end{table*}

\begin{table*}[t]
    \centering
    \caption{\textbf{Integrated comparison with the unified 3D-LLM block highlighted.}
    This table clarifies which rows provide broader literature context and which rows share the instance-centric unified 3D-LLM setting.}
    \label{tab:appendix_integrated_comparison}
    \vspace{0.2em}
    \begingroup

    \small
    \renewcommand{\arraystretch}{1.06}
    \setlength{\tabcolsep}{3.0pt}
    \begin{tabularx}{\textwidth}{@{}P{3.0cm}>{\raggedright\arraybackslash}X>{\raggedright\arraybackslash}X@{}}
        \toprule
        \textbf{Method} & \textbf{Input regime} & \textbf{Grounding interface} \\
        \midrule
        \rowcolor{referitsectionbg}\multicolumn{3}{@{}l@{}}{\textit{Specialized / task-specific grounding systems}}\\
        SeeGround & Mask3D proposals with rendered multi-view images and spatial text prompts & VLM scores candidate IDs over precomputed 3D detections using rendered-view evidence \\
        \rowcolor{referitsectionbg}\multicolumn{3}{@{}l@{}}{\textit{Reasoning-centric / agentic large-model pipelines}}\\
        VLM-Grounder & Grounding DINO detections, SAM masks, and multi-view RGB-D projection cues & Grounds in 2D views, lifts SAM masks to 3D, then merges projected evidence \\
        SPAZER & ScanNet detections with rendered 2D/3D views and large-model spatial verification prompts & Uses an agentic screening loop to verify candidate objects through 2D/3D observations \\
        \rowcolor{referitsectionbg}\multicolumn{3}{@{}l@{}}{\textit{Unified 3D-LLMs}}\\
        Chat-Scene & Mask3D proposals with Uni3D object features and DINOv2 multi-view appearance & Single-step object-ID route selects a candidate token inside a unified 3D-LLM \\
        Grounded 3D-LLM & Mask3D proposals encoded by the unified instance-centric 3D-LLM input pipeline & Single pointer-style \emph{QPG} route selects a query-index token for the target candidate \\
        \rowcolor{referitreferencebg}
        \textbf{SSR3D-LLM (ours)} & \textbf{Mask3D proposals with rotated-box geometry and DINOv2 multi-view appearance as proposal-level feature enhancement} & \textbf{LLM writes latent spatial reasoning steps; a geometry-aware scorer ranks candidates step by step} \\
        \bottomrule
    \end{tabularx}
    \endgroup
\end{table*}

\paragraph{Sr3D split behavior.}
Table~\ref{tab:appendix_sr3d_breakdown} restores the detailed Sr3D breakdown omitted from the main text for space.
Within the unified 3D-LLM family, SSR3D-LLM is strongest across the overall, Easy/Hard, and View-Dependent/View-Independent splits.
This is consistent with the intended role of latent steps: they give spatial and contextual cues a structured route into candidate ranking instead of relying on a single object readout.

\begin{table*}[t]
    \small
    \centering
    \caption{\textbf{Sr3D candidate-set breakdown.}
    This table restores the detailed Sr3D Top-1 accuracy breakdown from the earlier full ReferIt3D table.
    Rows without reported Sr3D breakdowns are omitted; bold marks the best result within the Unified 3D-LLMs.}
    \label{tab:appendix_sr3d_breakdown}
    \vspace{0.2em}
    \begingroup
    \small
    \setlength{\tabcolsep}{4.0pt}
    \renewcommand{\arraystretch}{1.05}
    \begin{tabularx}{0.92\textwidth}{@{}P{4.2cm}*{5}{Y}@{}}
        \toprule
        \textbf{Method} & \textbf{All} & \textbf{Easy} & \textbf{Hard} & \textbf{V-Dep.} & \textbf{V-Ind.} \\
        \midrule
        \rowcolor{referitsectionbg}\multicolumn{6}{@{}l@{}}{\textit{Specialized / task-specific grounding systems}}\\
        ReferIt3DNet (ECCV'20) & 40.8 & 44.7 & 31.5 & 39.2 & 40.8 \\
        MiKASA (CVPR'24) & 75.2 & 78.6 & 67.3 & 70.4 & 75.4 \\
        \midrule
        \rowcolor{referitsectionbg}\multicolumn{6}{@{}l@{}}{\textit{Unified 3D-LLMs}}\\
        Grounded 3D-LLM (arXiv'24) & 35.5 & 38.8 & 27.8 & 34.4 & 35.6 \\
        Chat-Scene (NeurIPS'24) & 31.5 & 34.3 & 25.1 & 33.6 & 31.4 \\
        \rowcolor{referitreferencebg}
        \textbf{SSR3D-LLM (ours)} & \textbf{69.3} & \textbf{74.4} & \textbf{57.1} & \textbf{53.3} & \textbf{70.0} \\
        \bottomrule
    \end{tabularx}
    \endgroup
\end{table*}

\paragraph{Cross-backbone transfer.}
Table~\ref{tab:appendix_transfer_chat_scene} is a paired-subset transfer diagnostic on Chat-Scene, using examples where annotations, proposals, and object-ID mappings can be aligned sample by sample.
It complements the official full-set results in the main paper and shows that the S3G grounding route can improve another unified 3D-LLM backbone when the candidate mapping is available.

\begin{table*}[t]
    \centering
    \caption{\textbf{Cross-model transfer evidence on Chat-Scene.}
    This subset uses samples for which annotations, proposal outputs, and object-ID mappings can be paired, and complements the official full-set results reported in the main tables.
    It measures cross-backbone adaptation on the paired evaluation subset.}
    \label{tab:appendix_transfer_chat_scene}
    \vspace{0.2em}
    \begingroup
    \small
    \renewcommand{\arraystretch}{1.05}
    \begin{tabularx}{0.88\textwidth}{@{}P{6.2cm}*{2}{Y}@{}}
        \toprule
        \textbf{Method} & \textbf{Nr3D Overall} & \textbf{Sr3D Overall} \\
        \midrule
        Chat-Scene & 25.9 & 31.5 \\
        Chat-Scene + SSR grounding & 30.3 & 40.5 \\
        \bottomrule
    \end{tabularx}
    \endgroup
\end{table*}

\paragraph{One-pass trace efficiency.}
Before adopting reserved step markers, we also tested an autoregressive trace prototype inspired by referential-chain supervision: the LLM first decodes an ordered textual referential chain, and the decoded chain is then used to obtain step representations for candidate scoring.
This prototype produced comparable grounding behavior in early experiments, but it required token-by-token generation at inference.
Table~\ref{tab:chainlen_efficiency} therefore reports the runtime gap between this AR trace route and the final one-pass latent-step route.
The final model keeps the trace latent: all step representations are read from reserved markers in one LLM forward, and the inspectable trace comes from step-wise candidate scores rather than generated text.

\begin{table*}[t]
    \centering
    \caption{\textbf{Runtime for step-length control.}
    We report wall-clock time and throughput (utterances/second) on a single GPU.
    One-pass uses one LLM forward with a latent length predictor; AR trace decoding generates up to 64 new tokens.}
    \label{tab:chainlen_efficiency}
    \vspace{0.2em}
    \begingroup
    \small
    \setlength{\tabcolsep}{4.0pt}
    \renewcommand{\arraystretch}{1.08}
    \begin{tabularx}{0.92\textwidth}{@{}XYYYYY@{}}
        \toprule
        \multirow{2}{*}{\textbf{Dataset}} &
        \multicolumn{2}{c}{\textbf{One-pass}} &
        \multicolumn{2}{c}{\textbf{AR trace}} &
        \multirow{2}{*}{\textbf{Speedup}} \\
        \cmidrule(lr){2-3}\cmidrule(lr){4-5}
        & \textbf{Time (s)} & \textbf{Utts/s} & \textbf{Time (s)} & \textbf{Utts/s} & \textbf{AR/one-pass} \\
        \midrule
        Nr3D & 94.1 & 21.8 & 314.5 & 3.2 & 6.8$\times$ \\
        Sr3D & 96.5 & 21.2 & 323.0 & 3.1 & 6.8$\times$ \\
        \bottomrule
    \end{tabularx}
    \endgroup
\end{table*}

\subsection{Capability preservation examples}
\label{sec:appendix_cap_examples}
Quantitative scores summarize average behavior, but a 3D-LLM should also \emph{look} normal on everyday language outputs.
We therefore include qualitative examples for non-grounding tasks (without \texttt{\textless geom\textgreater}) under the same checkpoint discussed in Section~\ref{sec:exp_results}.
Table~\ref{tab:cap_preserve_examples} lists randomly sampled inputs and the corresponding generations from both the \emph{QPG} baseline and SSR3D-LLM, with a scene snapshot for context.
All examples are randomly sampled with a fixed seed.
These examples complement the quantitative preservation results in the main paper: when \texttt{\textless geom\textgreater} is absent, SSR3D-LLM still uses the default language pathway for ordinary dialog, QA, and caption-style outputs.

\newcolumntype{L}[1]{>{\raggedright\arraybackslash}m{#1}}
\newcolumntype{C}[1]{>{\centering\arraybackslash}m{#1}}
\newcommand{\capexcell}[1]{#1}
\newcommand{\capexsceneA}{%
{\centering
\includegraphics[width=\linewidth]{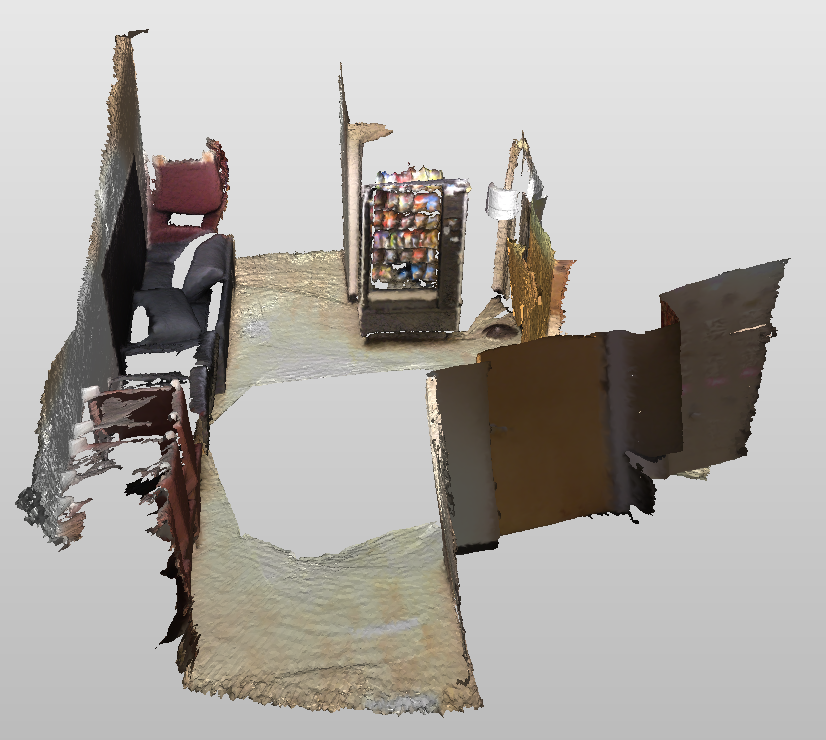}\\[-1pt]
{\scriptsize\texttt{scene0019\_01}}\par}%
}
\newcommand{\capexsceneB}{%
{\centering
\includegraphics[width=\linewidth]{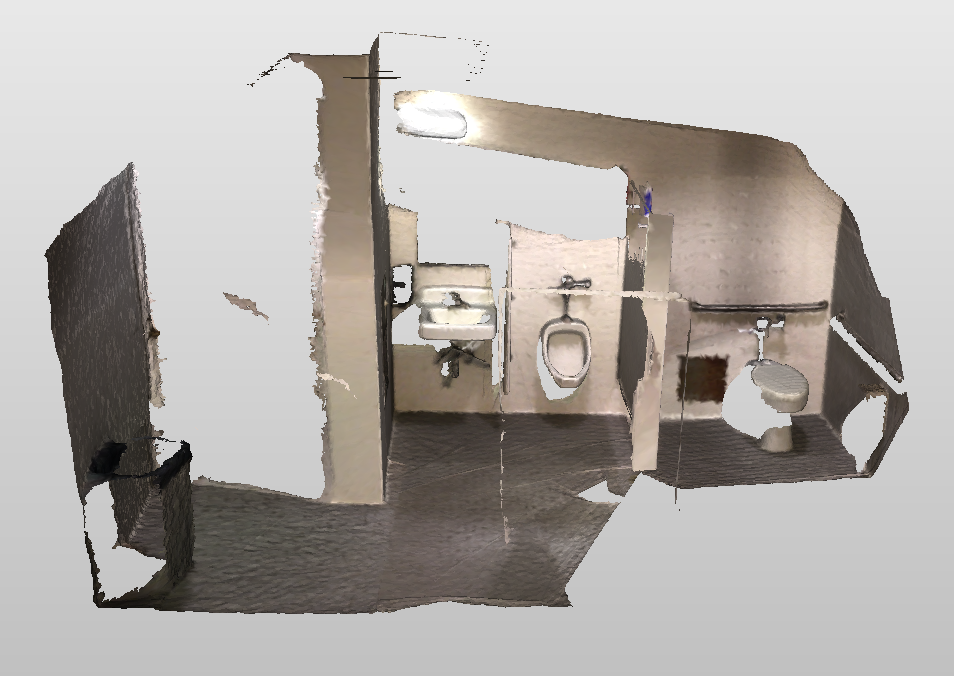}\\[-1pt]
{\scriptsize\texttt{scene0441\_00}}\par}%
}
\newcommand{\capexsceneC}{%
{\centering
\includegraphics[width=\linewidth]{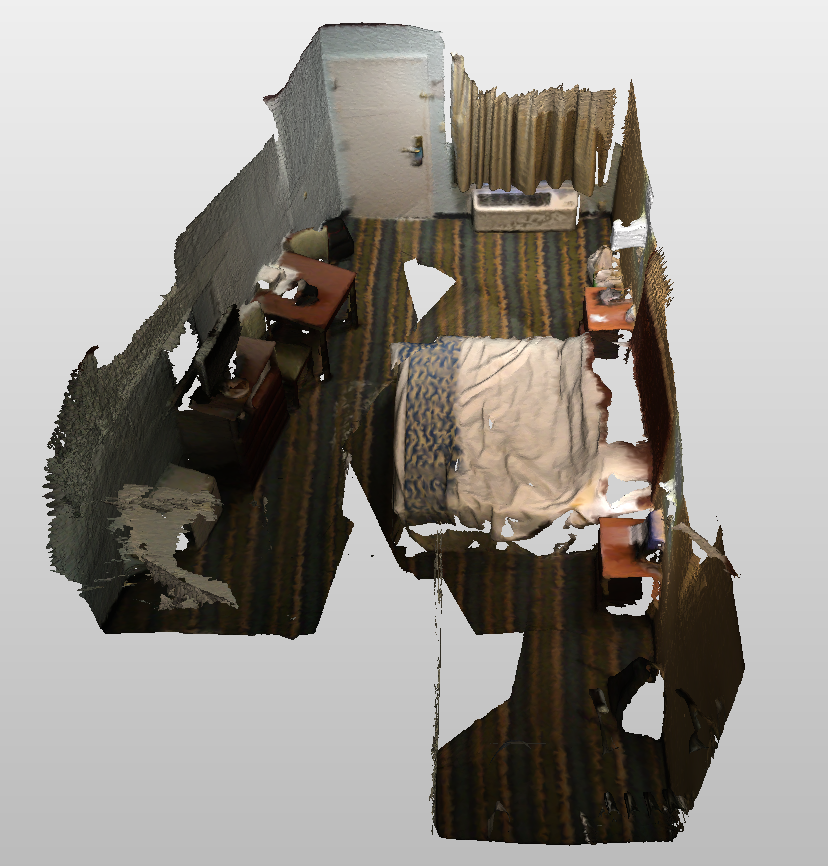}\\[-1pt]
{\scriptsize\texttt{scene0389\_00}}\par}%
}
\newcommand{\capexsceneD}{%
{\centering
\includegraphics[width=\linewidth]{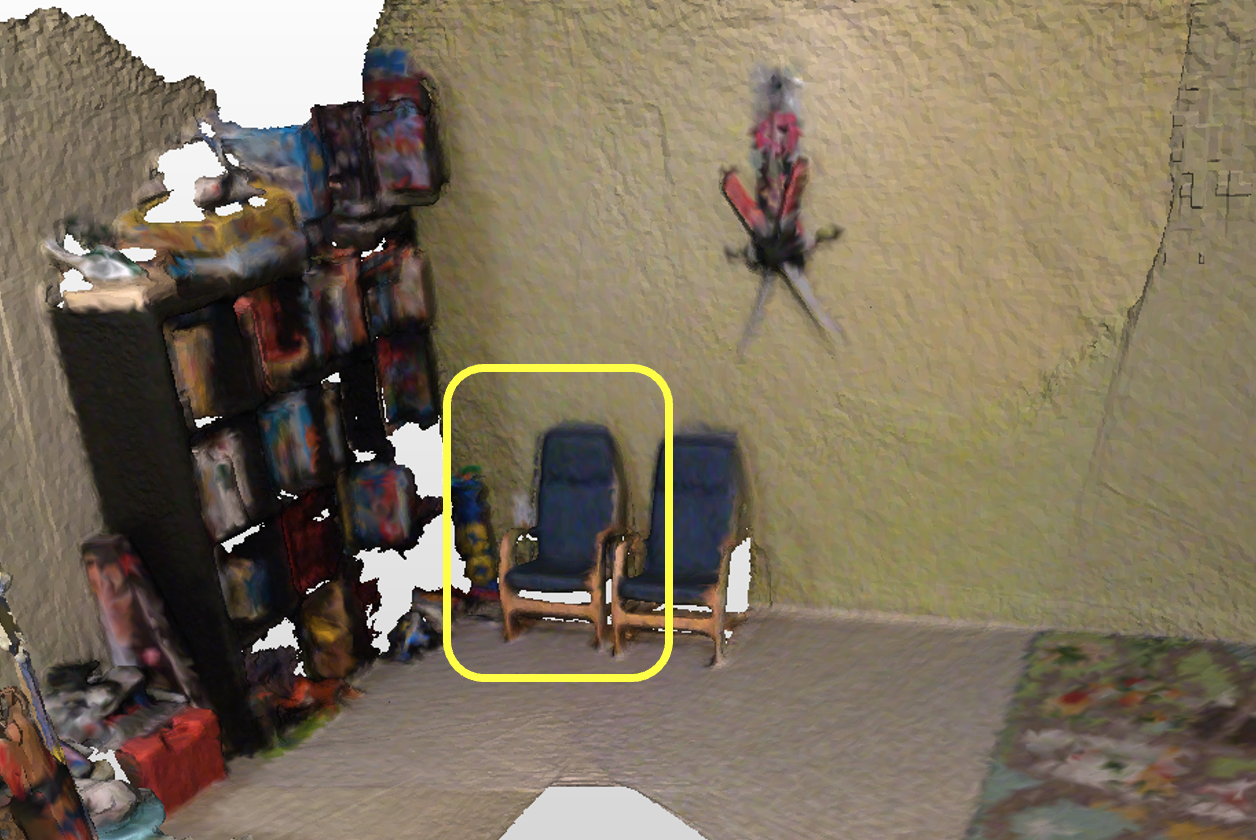}\\[-1pt]
{\scriptsize\texttt{scene0568\_01}}\par}%
}

\begin{table*}[t]
    \centering
    \caption{\textbf{Qualitative capability preservation (random samples).}
    Same checkpoint and the default language pathway (no \texttt{\textless geom\textgreater}).}
    \label{tab:cap_preserve_examples}
    \vspace{0.2em}
    \begingroup
    \small
    \setlength{\tabcolsep}{2.5pt}
    \renewcommand{\arraystretch}{1.05}
    \begin{tabular*}{\textwidth}{@{\extracolsep{\fill}}C{0.12\textwidth}L{0.085\textwidth}L{0.205\textwidth}L{0.26\textwidth}L{0.26\textwidth}}
        \toprule
        \textbf{Scene} & \textbf{Task} & \textbf{Input} & \textbf{\emph{QPG} baseline output} & \textbf{SSR3D-LLM output} \\
        \midrule
        \capexsceneA & \capexcell{Dialog} & \capexcell{Can you describe the general layout of this setting?} & \capexcell{Armchairs of curved backrest chairs have various features like leather and metallic legs.} & \capexcell{Certainly! Of curved backrest chairs have black upholstery and metallic legs.} \\
        \capexsceneB & \capexcell{ScanQA} & \capexcell{To the left of the urinal is what? Please answer with a single word or phrase.} & \capexcell{toilet paper dispenser} & \capexcell{toilet paper dispenser} \\
        \capexsceneC & \capexcell{Scan2Cap} & \capexcell{Describe the aesthetics of the object and its orientation within its immediate vicinity.} & \capexcell{there is a rectangular picture; it is on the wall above the bed.} & \capexcell{there is a rectangular picture; it is on the wall above the bed.} \\
        \capexsceneD & \capexcell{ObjDesc} & \capexcell{Could you give us a quick overview of the object?} & \capexcell{A wooden chair with a vertical slatted back design, upholstered in light-colored fabric.} & \capexcell{A black office chair with a mesh backrest.} \\
        \bottomrule
    \end{tabular*}
    \endgroup
\end{table*}

\subsection{Reproducibility, Assets, and Responsible Use}
\label{sec:appendix_repro_assets_responsible}

\paragraph{Reproducibility.}
The main experiments use ReferIt3D/Nr3D/Sr3D, ScanRefer, and Multi3DRef with the official splits and metrics described in Section~\ref{sec:exp_benchmarks}.
Main-table grounding results use Mask3D proposal-based candidates, proposal matching or instance-to-candidate mapping, and fixed evaluation scripts so that methods are evaluated under the same candidate hit criterion within each benchmark protocol.
The release package will include the training, evaluation, referential-cue preprocessing, mapping, and table-generation scripts used for the reported results, together with configuration files and checkpoint files or checkpoint-reconstruction instructions.
For external benchmarks and pretrained models, users should obtain the original assets from their official sources and follow the corresponding terms of use.

\begin{figure}[t]
  \centering
  \makebox[\textwidth][c]{\includegraphics[width=1.08\textwidth]{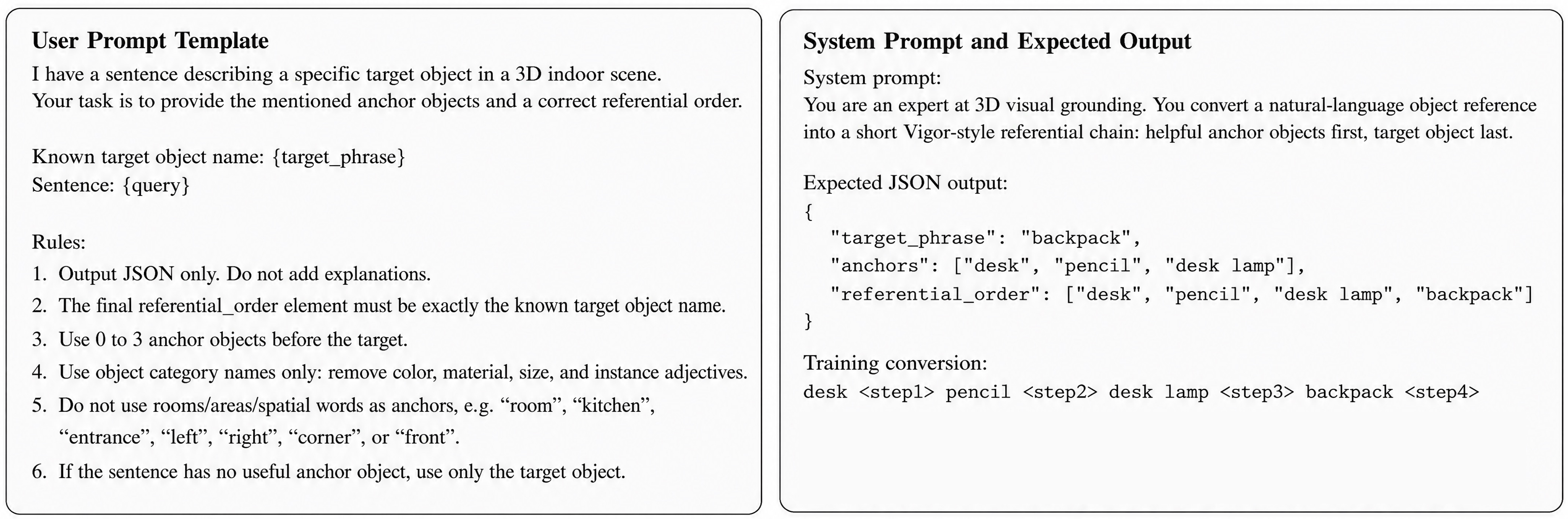}}
  \caption{\textbf{Offline referential-cue annotation prompt.}
  The Qwen/vLLM annotator receives the target phrase and query, returns JSON anchors plus a target-last referential order, and the preprocessing script converts that order into reserved step tokens.
  These generated cues supervise latent slots during training; inference uses only the query and Mask3D proposal representation, not generated or human-written chains.}
  \label{fig:referential_cue_prompt}
  \vspace{-0.5em}
\end{figure}

\paragraph{Training and experimental details.}
SSR3D-LLM uses a Tiny-Vicuna-1B unified 3D-LLM backbone, Mask3D proposals, \(K{=}4\) latent spatial reasoning steps, \(M{=}16\) memory tokens, rotated-box geometry, optional DINOv2 proposal appearance, and the target-class head/pred-class filter where reported.
Training follows the three-stage protocol in Section~\ref{sec:method_training}: scorer warm-start, latent-step workspace training, and continuation with the inference-time prompt.
For the ReferIt3D latent-step model in Table~\ref{tab:referit_main}, the one-pass base training uses Phase A for 20 epochs with batch size 64 and learning rate \(1{\times}10^{-4}\), then Phase B for 50 epochs with batch size 32 and learning rate \(5{\times}10^{-5}\).
Phase A keeps the listener fixed and uses referential-cue distillation; Phase B trains the listener and uses a language-class auxiliary weight of \(0.05\).
Figure~\ref{fig:referential_cue_prompt} shows the offline Qwen/vLLM prompt template used to generate these referential cues: it requests object-category anchors and a target-last referential order in JSON, then converts the order into reserved step tokens for training-time latent-slot supervision only.
The Stage-C continuation removes step phrases from the prompt and trains the reserved step markers from the query for 50 epochs with batch size 64 and learning rate \(1{\times}10^{-4}\).
The LLM-side adapter applies LoRA to the last four transformer layers' \(q\)- and \(v\)-projection matrices with rank 8, alpha 16, and dropout 0.0; the base Tiny-Vicuna weights remain frozen in these stages.
For language-capability preservation, we keep the same multi-task SFT recipe as Grounded 3D-LLM over dialog, ScanQA, Scan2Cap, and object-description language data; inputs without \texttt{\textless geom\textgreater} are trained and evaluated through the default generation pathway.
For ScanRefer and Multi3DRef, we continue from the Stage-C checkpoint with trainable listener and LLM-side adapter/projection parameters, initial learning rate \(1{\times}10^{-5}\), listener learning rate \(1{\times}10^{-5}\), LLM-side learning rate \(2{\times}10^{-5}\), and evaluation after every epoch.
The selected ScanRefer run uses batch size 48, DINOv2 appearance weight \(\alpha{=}1.0\), and target-class loss weight \(0.2\); the selected Multi3DRef run uses batch size 40 and DINOv2 appearance weight \(\alpha{=}2.0\).
The released configuration files specify the remaining command-level settings, optimizer choices, schedules, seeds, checkpoint paths, and run-specific overrides for each reported result.

\paragraph{Compute resources.}
Training and evaluation were run on NVIDIA A100 80GB GPU workers; the final grounding adaptations and the runtime measurements use single-GPU execution.
Table~\ref{tab:runtime_matched} reports post-proposal grounding-readout latency, throughput, and peak memory on a single GPU while excluding Mask3D proposal extraction; Appendix Table~\ref{tab:chainlen_efficiency} reports additional one-pass trace runtime.
Because exact hardware allocation and total GPU-hours vary across reruns and baselines, the release includes command-level logs, device metadata when available, and configuration files for compute accounting rather than a single aggregate compute number.

\paragraph{Grounding readout runtime protocol.}
Table~\ref{tab:runtime_matched} is a decision-stage diagnostic after proposal-level candidate inputs have already been prepared.
Timing starts after batch tensors are available on GPU and covers the forward path that converts object/query representations into the final grounding prediction, with CUDA synchronization enabled before recording wall-clock time.
All rows use batch size 1, no dataloader workers, 20 warm-up queries, and up to 500 measured queries per dataset/method; peak memory is recorded with \texttt{torch.cuda.max\_memory\_allocated}.
For Grounded 3D-LLM, the hook measures the \emph{QPG} grounding readout; for Chat-Scene, it measures the evaluation forward pass \texttt{model(**batch, is\_eval=True)}; for SSR3D-LLM, it measures the one-pass S3G listener/scorer readout.
The measurement excludes ScanNet preprocessing, Mask3D/CLASP proposal generation, proposal-feature preprocessing, DINOv2 feature extraction, referential-cue annotation, dataloader I/O, and CPU-side preprocessing.
Grounded 3D-LLM/SSR3D-LLM and Chat-Scene were run on separate NVIDIA A100 80GB GPU workers; Table~\ref{tab:runtime_matched} therefore reports readout-stage efficiency under the measured setups, while end-to-end runtime also depends on the shared upstream perception and preprocessing pipeline.
The released logs and scripts record the exact commands, warm-up and measurement counts, speed hooks, output directories, and device metadata available for each run.

\paragraph{Open assets.}
New released assets include SSR3D-LLM training/evaluation code, S3G modules and configuration files, referential-cue annotations or scripts to regenerate them, evaluation mapping utilities, and visualization scripts for trace and qualitative figures.
We release derived annotations, code, and configs as allowed, while raw datasets, pretrained backbones, Mask3D assets, DINOv2 models, and LLM weights remain governed by their original licenses and access terms.
Release documentation lists required external assets and where to obtain them.

\paragraph{Broader impacts.}
SSR3D-LLM may benefit embodied AI, navigation, assistive robotics, and spatial scene understanding by improving fine-grained grounding of objects and spatial relations in 3D scenes.
The same capability can inherit biases and errors from indoor datasets, upstream proposal detectors, and pretrained visual/language models.
Incorrect grounding in embodied systems could lead to unsafe or disruptive actions if deployed without validation.
This paper evaluates static ScanNet-style research benchmarks and does not deploy a robot or safety-critical system; deployment with richer visual front ends should include task-specific validation, monitoring, and human oversight.

\subsection{Trace Diagnostics and Error Analysis}
\label{sec:appendix_trace_diagnostics}

The following figures analyze saved per-step scores, masks, and confidence values from the evaluated checkpoints.
They provide diagnostic views of how latent steps affect ranking dynamics, where high-confidence errors occur, and how the \emph{QPG} baseline behaves under the same candidate-set analysis.

\paragraph{Example traces.}
Figure~\ref{fig:trace_score_heatmaps} visualizes individual step-wise score trajectories.
These examples complement the main qualitative case by showing the same information in a compact heatmap form: which candidates become competitive, when the ground-truth object rises, and how much probability mass remains outside the tracked candidates.

\IfFileExists{figs/trace_score_heatmaps.pdf}{
\begin{figure*}[t]
    \centering
    \includegraphics[width=\textwidth]{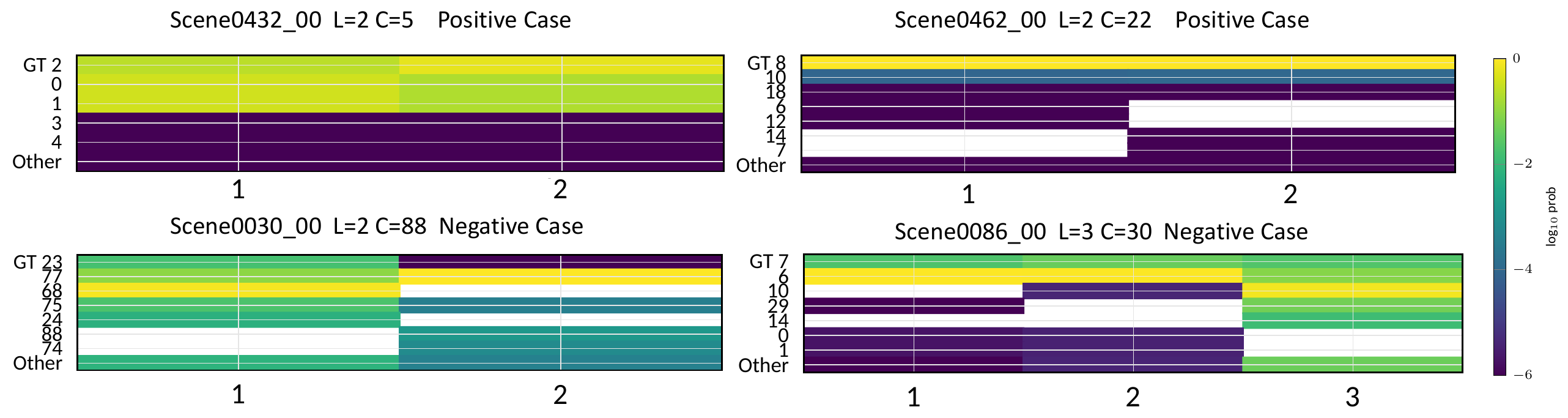}
    \caption{\textbf{Step-wise score heatmaps (examples).}
    For each query, we track the ground-truth (GT) object and the union of candidates that enter the per-step top-\(k\) list, and visualize their step-wise probabilities over effective steps (\(k\le L\)) as a heatmap (color indicates \(\log_{10}\) probability).
    Rows show candidate instance ids (and labels when available); ``Other'' aggregates the remaining probability mass outside the tracked candidates.
    These examples make the intermediate scoring process directly inspectable.}
    \label{fig:trace_score_heatmaps}
    \vspace{-0.6em}
\end{figure*}
}{}

\paragraph{Step-length masking analysis.}
Figure~\ref{fig:freeze_stats} validates the masking mechanism used by the recursive scorer:
updates are active only for valid steps \(k\le L\), and become inert once \(k>L\) (drift-free padded computation).
This explains why fixed maximum step markers can support variable-length queries without letting unused steps perturb the final ranking.

\paragraph{GT rank flow.}
Figure~\ref{fig:gtrankflow} aggregates the rank of the ground-truth object over normalized latent-step progress.
It provides a complementary view to the per-example heatmaps: instead of showing one query, it summarizes whether the recursive updates tend to move the target object closer to the top of the candidate list across evaluation traces.

\paragraph{Full-evaluation macro trends.}
We summarize \emph{all} evaluation utterances using the per-candidate probability distributions exported by our evaluation pipeline.
Figure~\ref{fig:macro_heatmaps_full} shows accuracy and high-confidence conditional error as a function of oracle step length \(L\) (clipped to \(K=4\)) and candidate-set size.
Figure~\ref{fig:macro_error_modes_full} further visualizes the confidence separation and the high-confidence conditional error across \(L\).
These plots make the residual failure modes easier to read: longer cue chains and larger candidate sets remain harder, while the coverage and conditional-error panels separate confident predictions from reliable confident predictions.

\IfFileExists{figs/eval_macro_heatmaps_full_qpg.pdf}{
\paragraph{\emph{QPG} diagnostic macro trends.}
Figures~\ref{fig:macro_heatmaps_full_qpg}--\ref{fig:macro_highconf_risk_full_qpg} repeat the same binning as Fig.~\ref{fig:macro_heatmaps_full}, but use an exported candidate-context diagnostic for the single-readout \emph{QPG} baseline.
This diagnostic is separate from the main benchmark tables and is used to inspect how the \emph{QPG} readout behaves across step-length and candidate-set-size bins.
Because its confidence proxy is derived from similarity scores rather than a calibrated listener softmax, we use a fixed-coverage high-confidence definition (top-5\% by \(p_{\max}\), per seed) and report \( \Pr(\text{wrong}\mid \text{high-conf}) \).

\begin{figure*}[h]
    \centering
    \includegraphics[width=\textwidth]{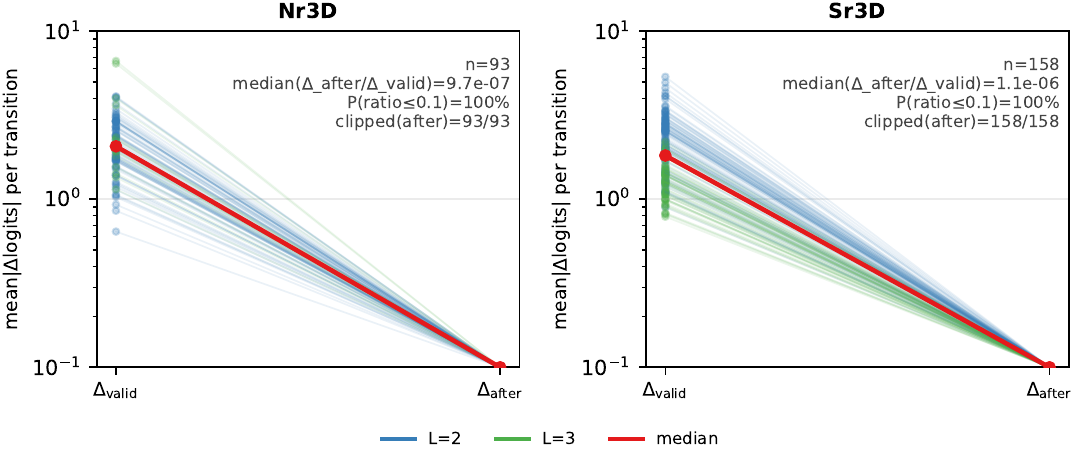}
    \caption{\textbf{Step-length masking analysis.}
    From exported stepwise traces, we directly compare the transition magnitude within valid steps (\(\Delta_{\mathrm{valid}}\)) and after termination (\(\Delta_{\mathrm{after}}\)).
    The near-zero \(\Delta_{\mathrm{after}}\) confirms step-length masking makes padded steps behaviorally inert.}
    \label{fig:freeze_stats}
    \vspace{+5em}
\end{figure*}

\IfFileExists{figs/figure4-202604281747.pdf}{
\begin{figure*}[ht]
    \centering
    \includegraphics[width=\textwidth]{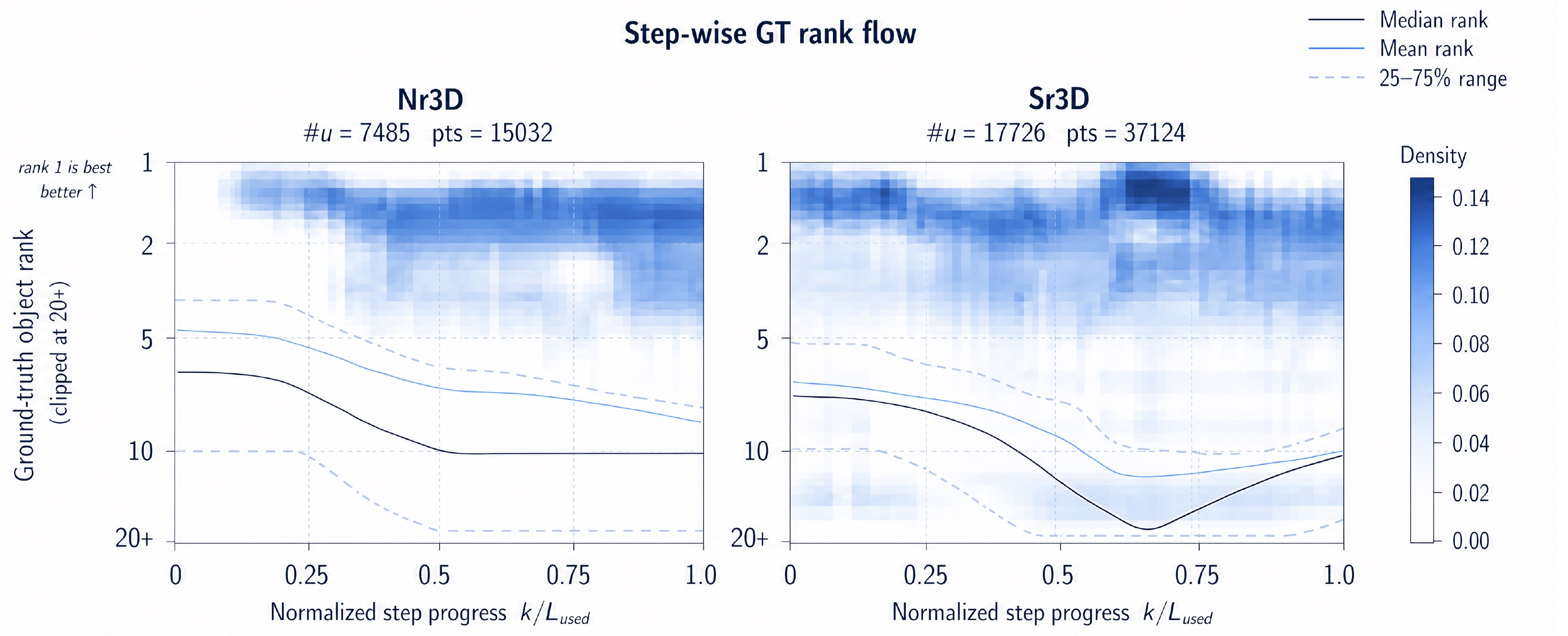}
    \caption{\textbf{GT rank flow from step-wise traces.}
    Smoothed density of GT rank over normalized step progress \(k/L_{\mathrm{used}}\).
    Columns are normalized within each progress step; curves summarize the median, mean, and 25--75\% interquartile range.
    This aggregate diagnostic shows whether recursive updates tend to move the ground-truth object toward the top of the candidate ranking.}
    \label{fig:gtrankflow}
    \vspace{-0.6em}
\end{figure*}
}{}

\begin{figure*}[h]
    \centering
    \includegraphics[width=\textwidth]{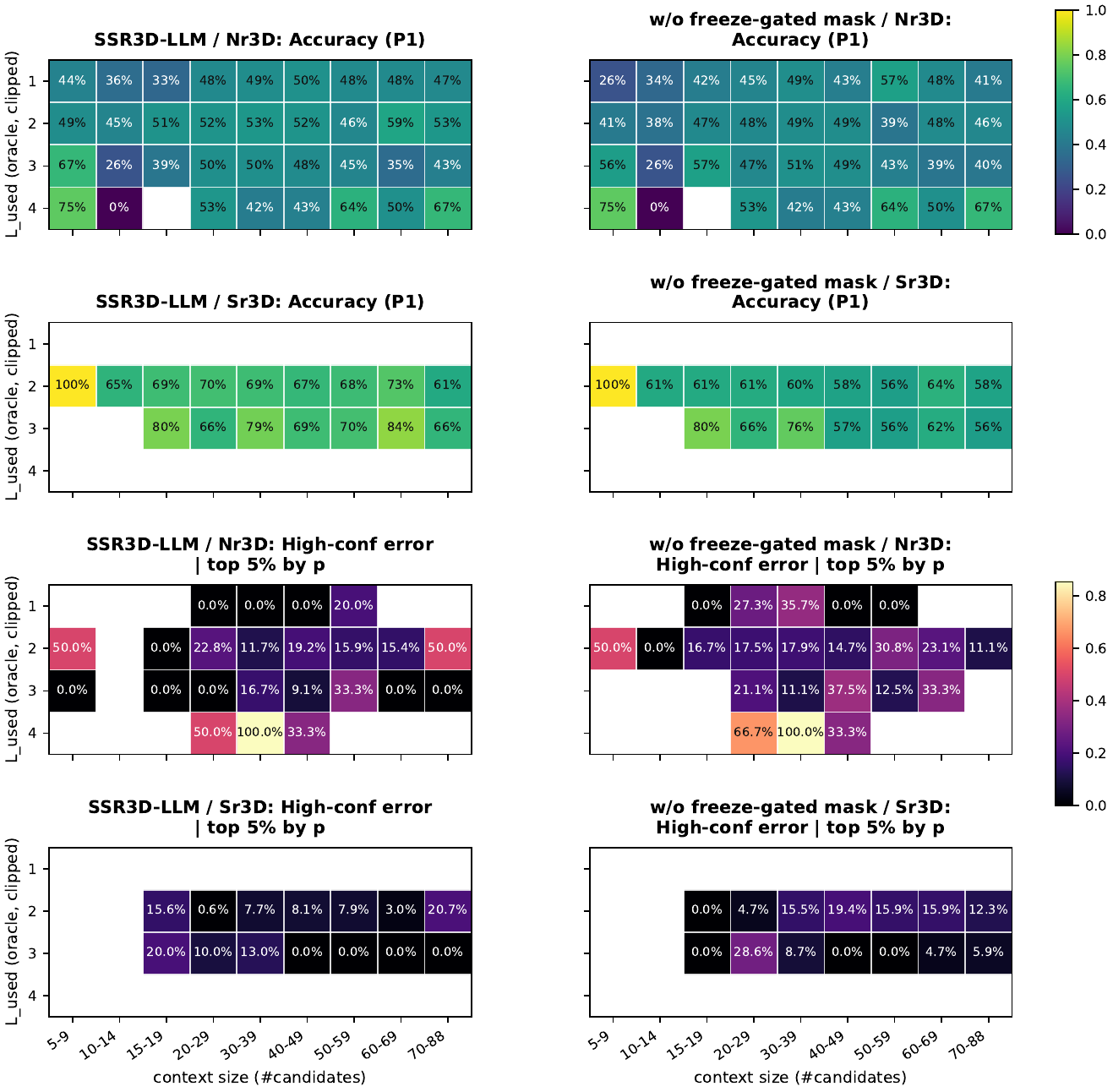}
    \caption{\textbf{Macro trends from full evaluation (Nr3D, Sr3D).}
    We aggregate the full evaluation set using the saved per-candidate probability distributions (no re-inference).
    Rows correspond to oracle step length \(L\) (from referential-cue annotations, clipped to \(K=4\)); columns bin the candidate-set size (\#objects).
    We show two methods side-by-side: SSR3D-LLM (ours) and a paired baseline \emph{w/o step-length mask} (fixed-\(K\) updates), using the same candidate sets and the same evaluation artifacts.
    Note Sr3D only contains \(L\in\{2,3\}\); rare Nr3D cases with \(L>4\) are clipped to \(4\) to match the model.
    The top row reports candidate-set accuracy; the bottom row reports \emph{high-confidence conditional error} (a.k.a.\ selective risk):
    we define \emph{high confidence} as the top-5\% predictions by \(p_{\max}\) (threshold computed per eval repeat), and report \( \Pr(\text{wrong}\mid \text{high-conf}) \) within each bin.
    Figure~\ref{fig:macro_highconf_coverage_full} reports the corresponding coverage \( \Pr(\text{high-conf}) \) (which can vary across bins), enabling correct interpretation of near-zero cells.}
    \label{fig:macro_heatmaps_full}
    \vspace{-0.6em}
\end{figure*}

\begin{figure*}[h]
    \centering
    \includegraphics[width=\textwidth]{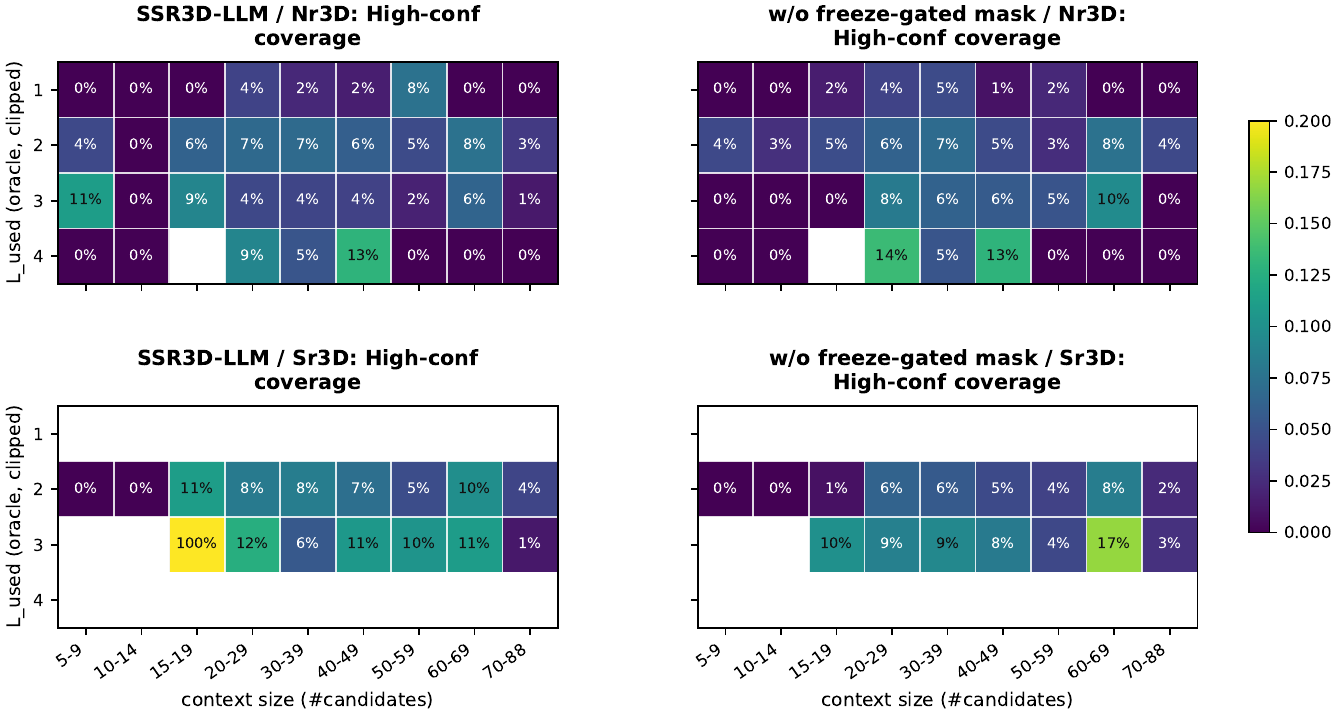}
    \caption{\textbf{High-confidence coverage from full evaluation.}
    Same binning as Fig.~\ref{fig:macro_heatmaps_full}.
    We report \( \Pr(\text{high-conf}) \), where \(\text{high-conf}\) is defined as top-5\% by \(p_{\max}\) (per eval repeat).
    These maps disambiguate whether a low conditional error cell stems from genuinely low risk or simply from low coverage.}
    \label{fig:macro_highconf_coverage_full}
    \vspace{-0.6em}
\end{figure*}

\begin{figure*}[t]
    \centering
    \includegraphics[width=\textwidth]{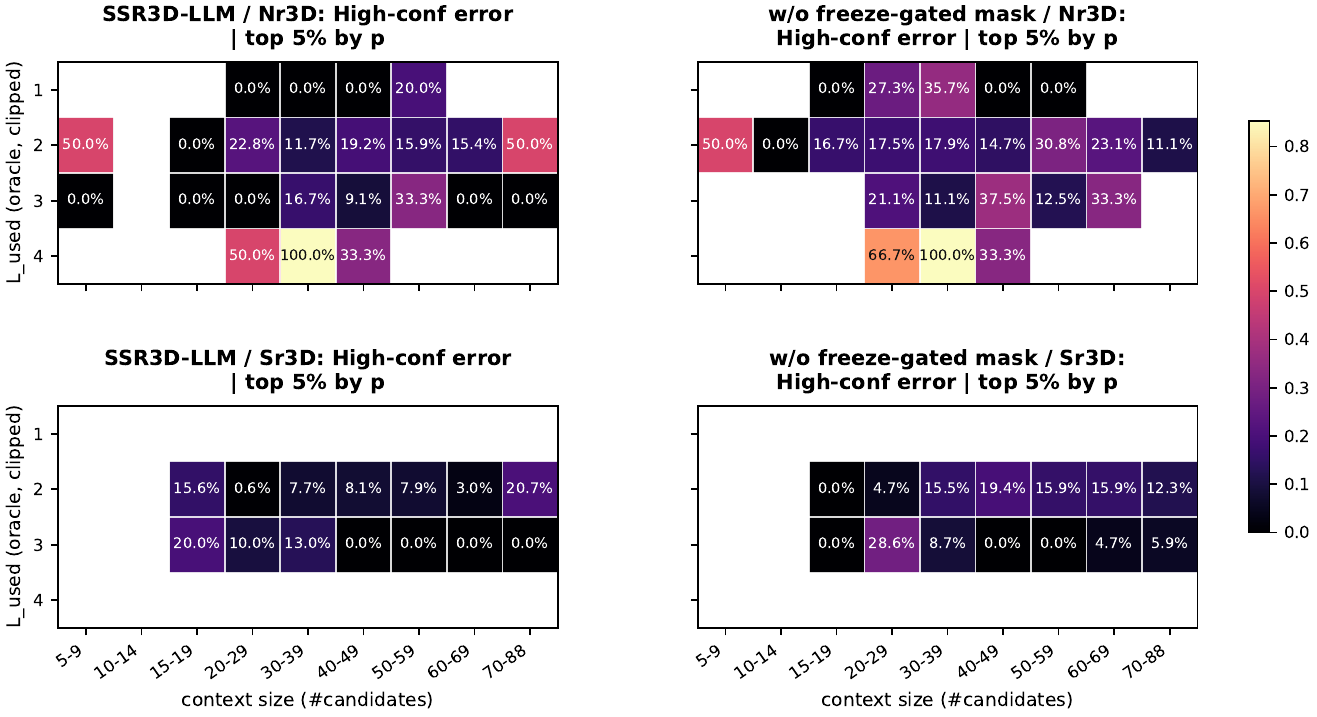}
    \caption{\textbf{High-confidence conditional error from full evaluation.}
    Same data as the bottom row of Fig.~\ref{fig:macro_heatmaps_full}, shown alone for readability.}
    \label{fig:macro_highconf_risk_full}
    \vspace{-0.6em}
\end{figure*}

\begin{figure*}[t]
    \centering
    \includegraphics[width=\textwidth]{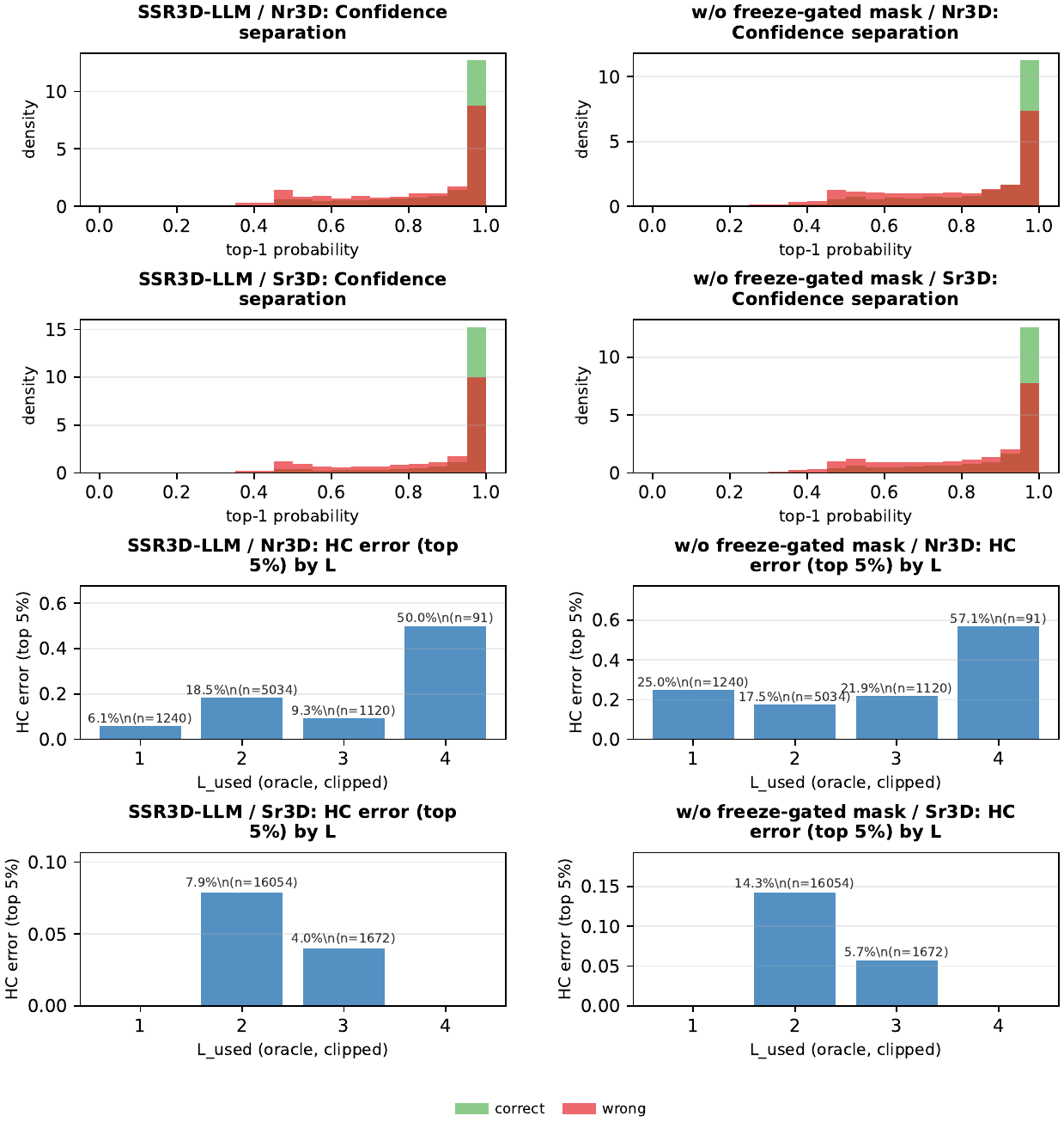}
    \caption{\textbf{Error-mode breakdown from full evaluation.}
    Top: top-1 probability histograms for correct vs wrong predictions.
    Bottom: high-confidence conditional error by oracle step length \(L\) (clipped to \(K=4\)) with sample counts.}
    \label{fig:macro_error_modes_full}
    \vspace{-0.6em}
\end{figure*}

\begin{figure*}[t]
    \centering
    \includegraphics[width=\textwidth]{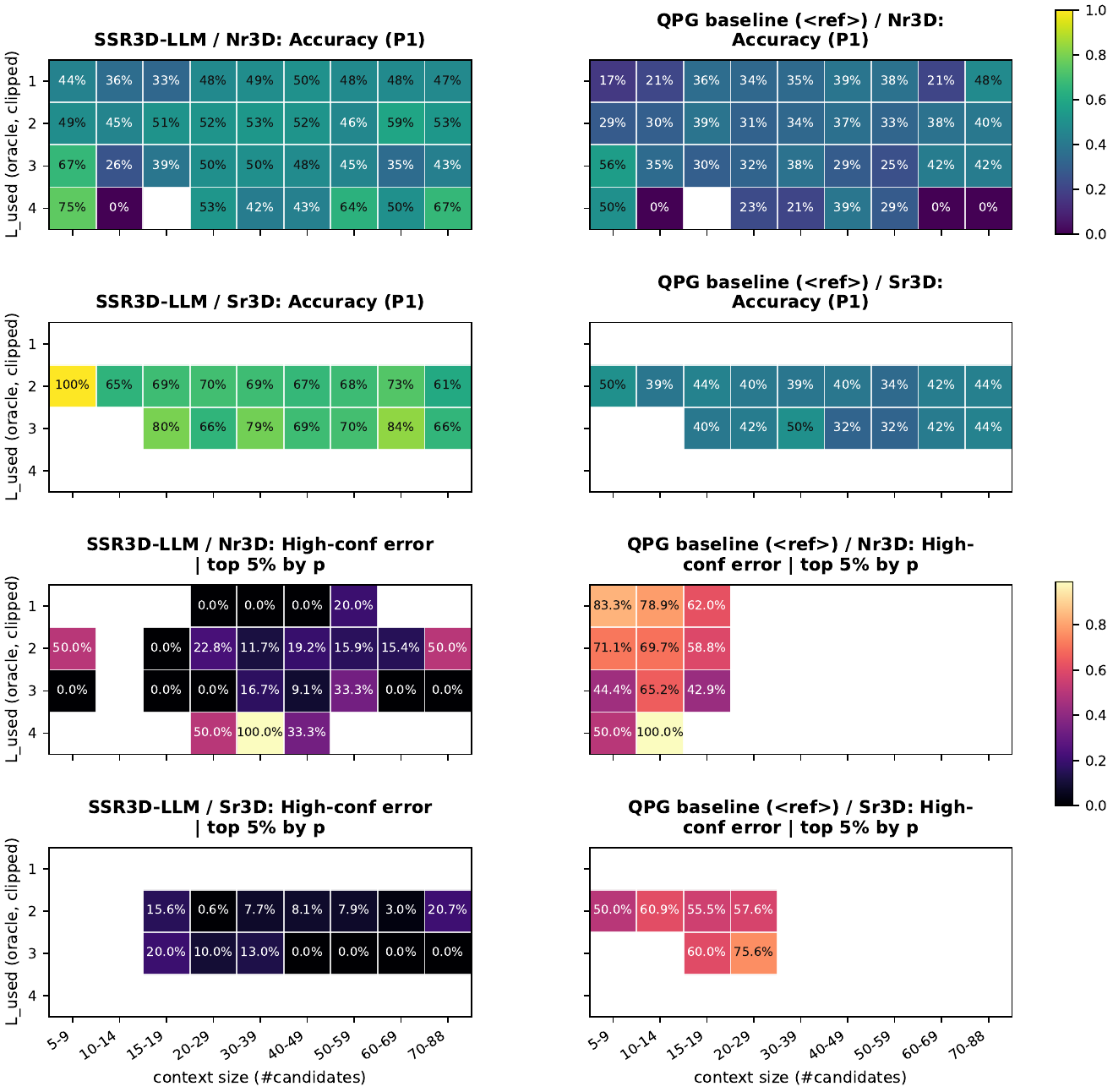}
    \caption{\textbf{\emph{QPG} diagnostic macro trends.}
    Same binning as Fig.~\ref{fig:macro_heatmaps_full}, with the right column showing a single-readout \emph{QPG} diagnostic on the exported candidate contexts.
    This figure is an appendix diagnostic rather than the main same-protocol benchmark comparison.
    The confidence proxy is a softmax-normalized top-1 probability among candidate-context scores (temperature \(T=1\)); because it comes from similarity scores rather than a calibrated listener softmax, high confidence is defined by fixed coverage: top-5\% by proxy \(p_{\max}\), computed per seed.
    Blank cells indicate bins with no samples or no high-confidence samples under this diagnostic.}
    \label{fig:macro_heatmaps_full_qpg}
    \vspace{-0.6em}
\end{figure*}
}{}

\IfFileExists{figs/eval_macro_error_modes_full_qpg.pdf}{
\begin{figure*}[t]
    \centering
    \includegraphics[width=\textwidth]{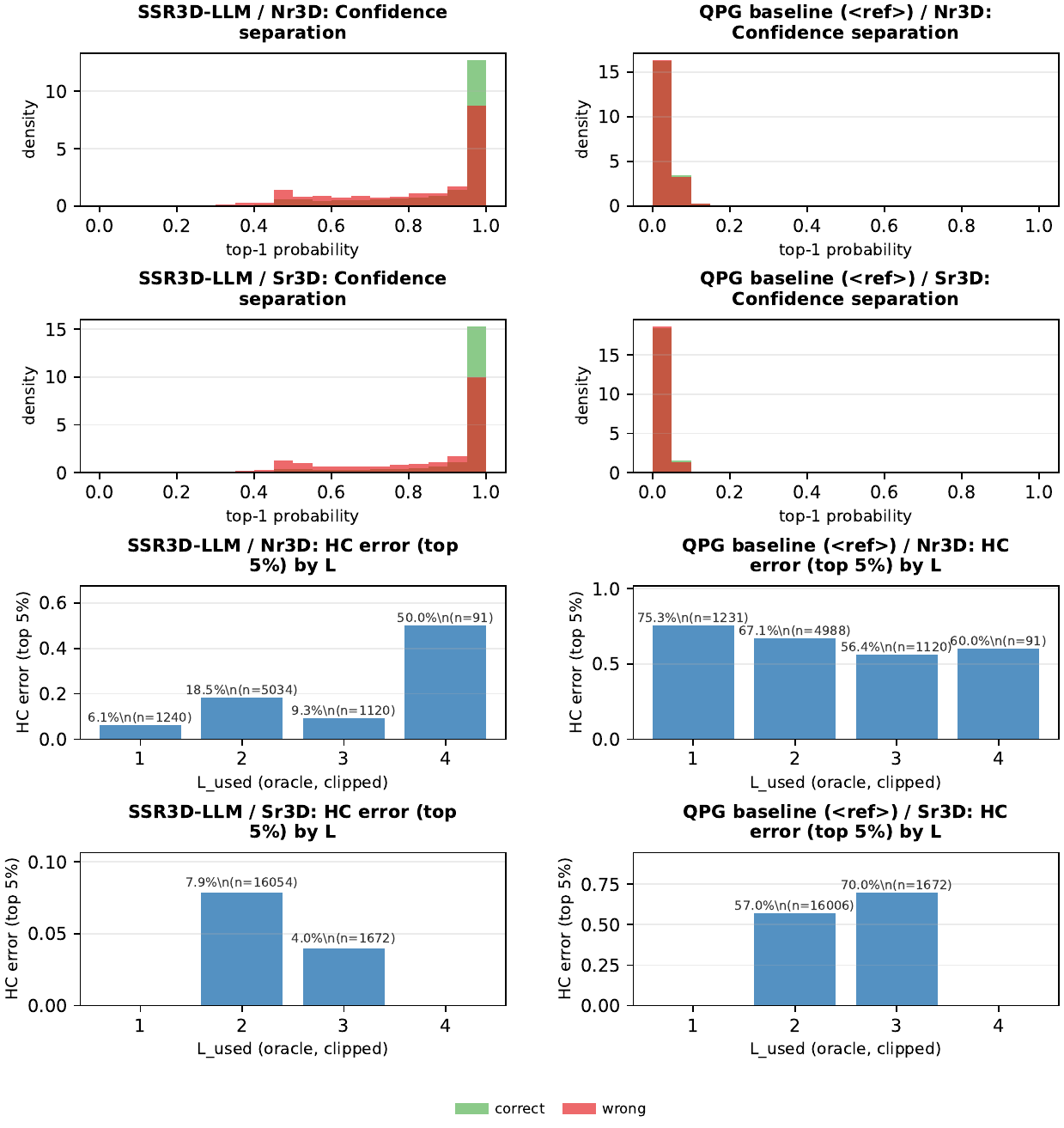}
    \caption{\textbf{\emph{QPG} diagnostic error-mode breakdown.}
    Same layout as Fig.~\ref{fig:macro_error_modes_full}, with the right column showing the exported candidate-context \emph{QPG} diagnostic.
    The high-confidence definition matches Fig.~\ref{fig:macro_heatmaps_full_qpg}: top-5\% by proxy \(p_{\max}\), computed per seed.}
    \label{fig:macro_error_modes_full_qpg}
    \vspace{-0.6em}
\end{figure*}
}{}

\IfFileExists{figs/eval_macro_highconf_coverage_full_qpg.pdf}{
\begin{figure*}[t]
    \centering
    \includegraphics[width=\textwidth]{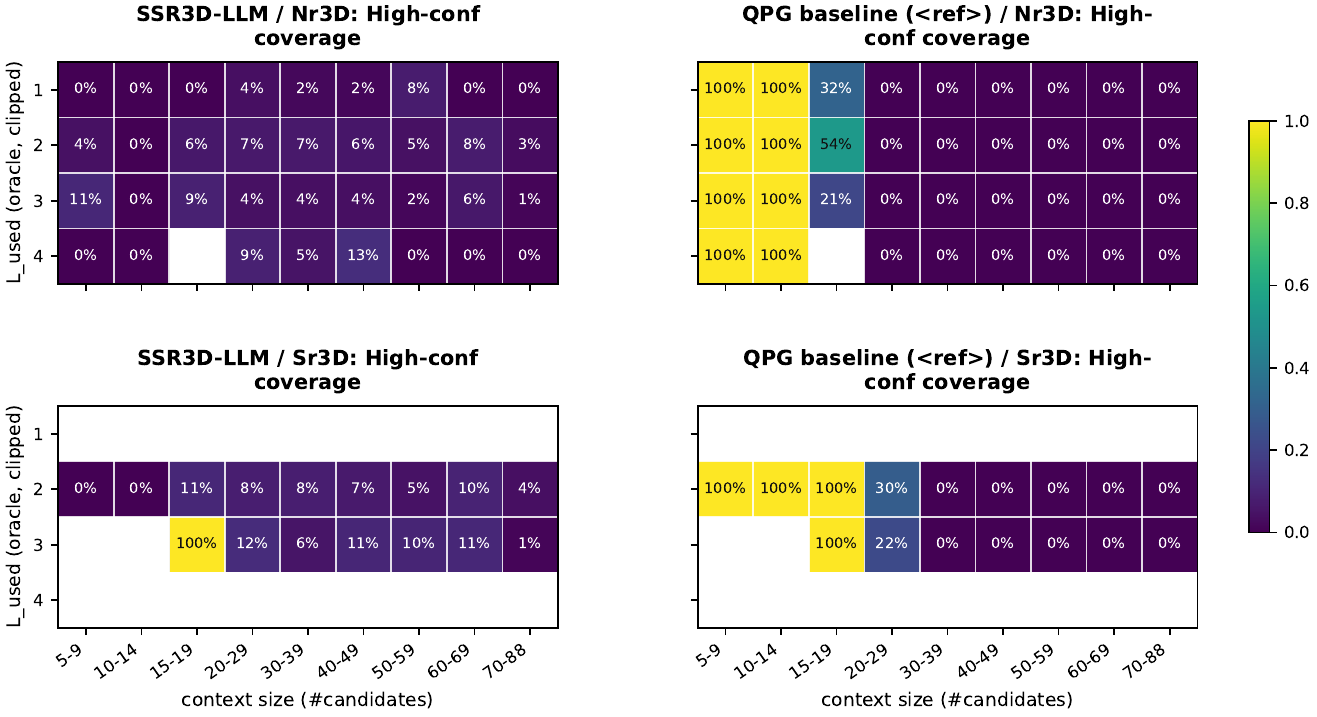}
    \caption{\textbf{\emph{QPG} diagnostic high-confidence coverage.}
    Same as Fig.~\ref{fig:macro_highconf_coverage_full}, with the exported candidate-context \emph{QPG} diagnostic column.}
    \label{fig:macro_highconf_coverage_full_qpg}
    \vspace{-0.6em}
\end{figure*}
}{}

\IfFileExists{figs/eval_macro_highconf_risk_full_qpg.pdf}{
\begin{figure*}[t]
    \centering
    \includegraphics[width=\textwidth]{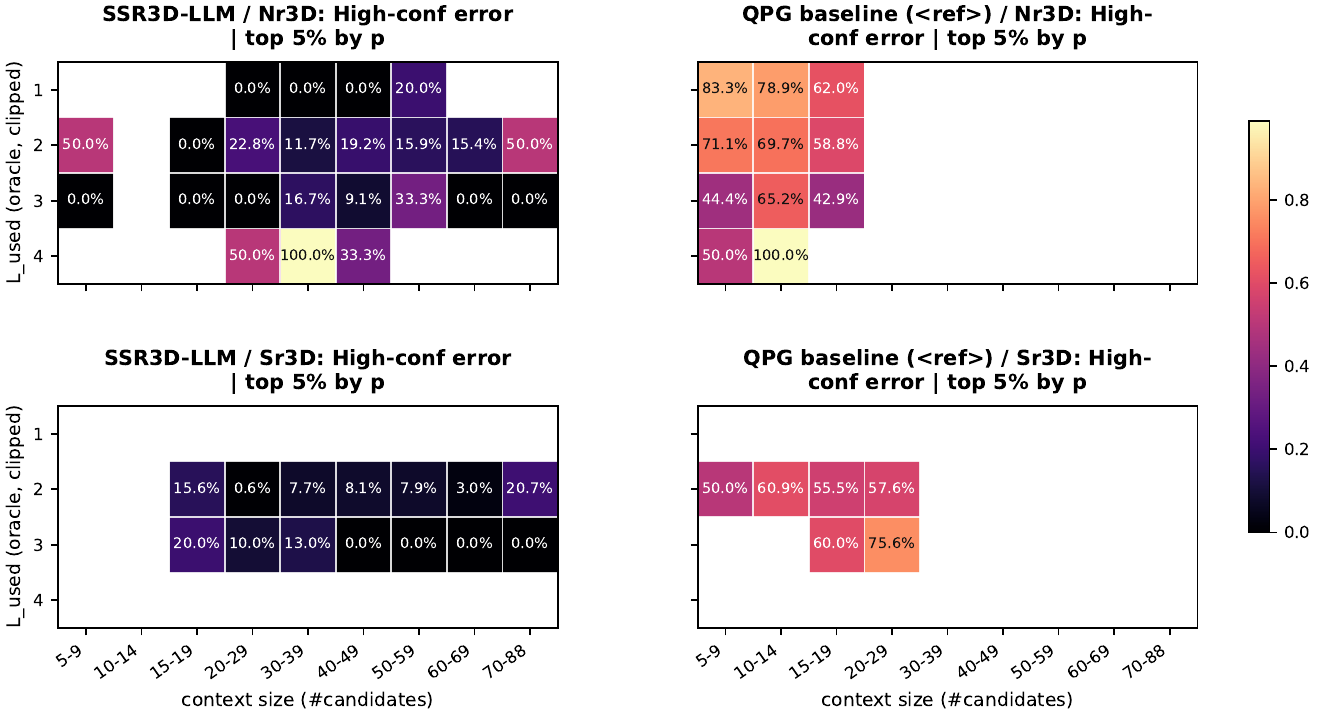}
    \caption{\textbf{\emph{QPG} diagnostic high-confidence conditional error.}
    Same as Fig.~\ref{fig:macro_highconf_risk_full}, with the exported candidate-context \emph{QPG} diagnostic column.}
    \label{fig:macro_highconf_risk_full_qpg}
    \vspace{-0.6em}
\end{figure*}
}{}

\clearpage

\end{document}